\documentclass[11pt]{article}
\pdfoutput=1
\usepackage{prior}

\usepackage{amsmath,amsfonts,bm}

\def\eqref#1{equation~\ref{#1}}

\def\1{\bm{1}}

\DeclareMathAlphabet{\mathsfit}{\encodingdefault}{\sfdefault}{m}{sl}
\SetMathAlphabet{\mathsfit}{bold}{\encodingdefault}{\sfdefault}{bx}{n}

\usepackage{placeins}
\tcbuselibrary{most}

\setlist{leftmargin=*,nosep}
\usepackage{etoolbox}
\makeatletter
\patchcmd{\prior@firstfoot}{\textsuperscript{\dag}Corresponding author. Email:}{Corresponding authors:}{}
  {\PackageWarning{paper}{could not pluralise the corresponding-author label}}
\makeatother

\title{When Less Memory Helps: Selective Recall and Compression Pressure in Recurrent Models}
\shorttitle{When Less Memory Helps}
\author{Core Francisco Park\textsuperscript{*$\dagger$}, Fuming Yang\textsuperscript{*}}
\authornote{\textsuperscript{*}Equal contribution. \textsuperscript{$\dagger$}Senior contribution.}
\correspondingauthor{Core Francisco Park, \texttt{core@priorcomputers.ai}; Fuming Yang, \texttt{fuming@priorcomputers.ai}}

\begin{document}

\begin{abstract}
While you read this abstract, try to remember how many points fine-tuning adds to cued recall. This instruction is already changing how your brain registers this abstract, and arguably is a major difference between recurrent models (RMs) with finite state and Transformers, which can attend back to any context. Consequently, an RM must commit to its finite memory what it thinks will be relevant in the future and compress information on the go. This intuition behind state and data dependent memory drove recent architectural designs like Gated DeltaNet and RWKV-7. However, we find that language-trained RMs do not actually behave this way. In a task we call cued key--value recall, a cue that names the future query helps mostly through exact matching: the models barely process its meaning, and telling them to ignore a record still helps them recall it. Fine-tuning adds 89 points to RWKV-7's cued recall (8\% to 97\% at 256 biographies), yet what the models learn remains largely token matching. Trained from scratch on a synthetic version of the task, RWKV-7 behaves rationally: uncued recall is limited by memory capacity, cued recall by length generalization, and a smaller state starts relying on the cue sooner. But the same limit that hurts recall can help learning: a finite state must compress, and we find that this \emph{inference-time compression pressure} can actually push a model towards learning better features and concepts. When trained only
to recall given pixels of a digit, an RM learns to draw the whole
digit, while a Transformer trained on the same data and loss simply copies
from its inputs. Finally, we show that these results compile into a phase diagram where
task load and memory load both contribute to inference-time pressure, creating
better-generalizing features. Now, do you remember how
many points fine-tuning added? Perhaps the cue helped, even though you do not have quadratic attention.
\end{abstract}

\maketitle
\section{Introduction}
\label{sec:intro}

\begin{figure}[t]
    \centering
    \includegraphics[width=\textwidth]{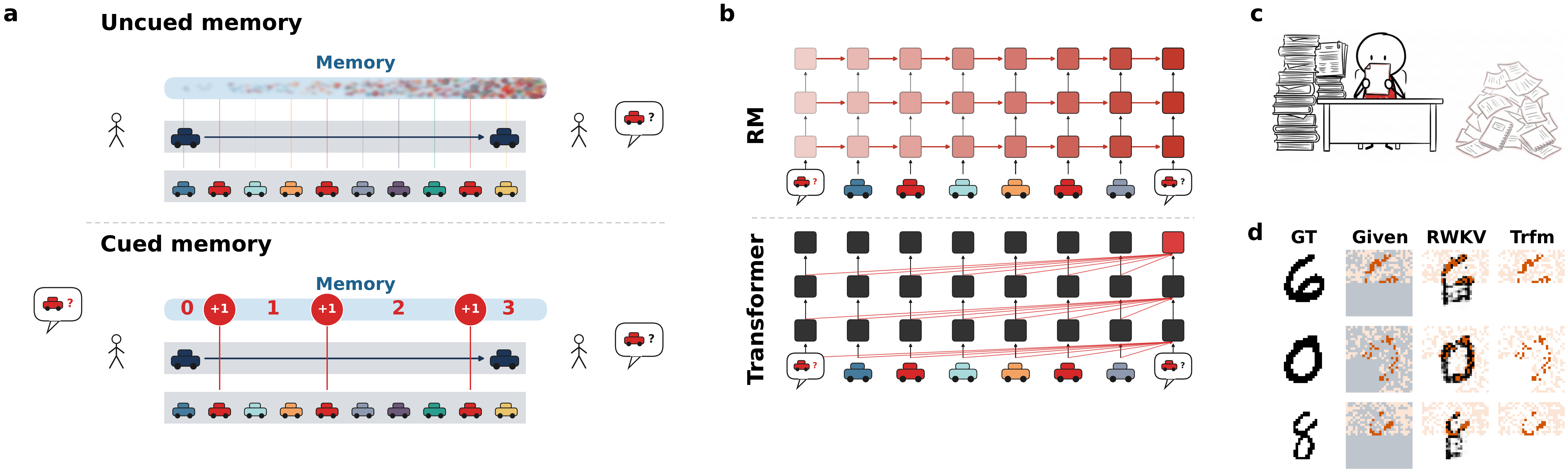}
    \caption{\textbf{A finite memory system must decide what to keep before the question arrives, and the same pressure can help learning.}
    \textbf{a.} A driver passes a stream of cars on the way to work. An uncued driver probably can't answer how many red cars they saw, but a driver who already know the question can easily perform the task.
    \textbf{b.} An RM carries a fixed-size state from token to token, so what it keeps is decided as each car arrives. A Transformer, however, can always attend back to every token and easily retrieve what it needs.
    \textbf{c.} A person reading 100 papers likely cannot remember even a single abstract perfectly. However, they end up with a highly compressed world model on the topic.
    \textbf{d.} We find \emph{inference-time compression pressure} in action when training sequence models to simply recall the pixels values they are given. Observed pixels are in orange and hidden pixels in gray. RWKV-7, an RM, fills in the hidden part of the digit, while a Transformer (Trfm) reproduces only the given pixels. The RWKV has learned a compressed concept while the Transformer simply recalls.}
    \label{fig:overview}
\end{figure}

Imagine driving to work while dozens of cars pass by (Fig.~\ref{fig:overview}a).
When you arrive, a co-worker asks how many of them were red.
If you are a human, you almost certainly cannot answer.
However, if your co-worker had asked the question before you left, the task would likely not be too hard for humans as well.

This example demonstrates that humans have evolved to save the most important information by letting their goals shape what they perceive and store.
Attending to one aspect of a scene can hide even a salient unexpected event \citep{Simons_1999}.
Simply telling people what to remember changes what they store: items followed by an instruction to remember are recalled better than items followed by an instruction to forget \citep{Hall_2021}, and people selectively remember the items they are told are worth more \citep{Castel_2002,Knowlton_2022}.

A recurrent model (RM)\footnote{We use the term \emph{recurrent model} (RM) for any sequence model with a fixed-size state: RNNs, LSTMs, GRUs, Linear Transformers, Mamba, DeltaNet, Gated DeltaNet, RWKV and their relatives.} is in the human's position (Fig.~\ref{fig:overview}b).
It must compress all useful incoming information into a finite state, and thus it must decide what to keep before it knows what will be asked.
A Transformer \citep{vaswani2023attentionneed} given a video of the drive has no such problem, since self-attention always allows it to look back at any part of the context.
This is arguably the single biggest difference between RMs and Transformers.

This difference drives much of RM architecture design.
Many recent RMs are built precisely so that the state can choose what to keep: input-dependent gates and decays decide what to forget \citep{peng2023rwkvreinventingrnnstransformer, gu2024mambalineartimesequencemodeling, de2024griffinmixinggatedlinear, yang2025gateddeltanetworksimproving, peng2025rwkv7gooseexpressivedynamic, kimiteam2025kimilinearexpressiveefficient}, the delta rule overwrites stale associations \citep{yang2025parallelizinglineartransformersdelta}, and test-time learning trains the state as a memory on the fly \citep{sun2025learninglearntesttime, behrouz2024titanslearningmemorizetest, behrouz2025nestedlearningillusiondeep}.
In other words, how to selectively save and forget is at the core of these architectures' design.
However, outside the minds of their designers, these architectures are evaluated like any other Transformer: trained on standard data and scored on scaling curves, throughput and benchmarks.

\textbf{This work.} This leaves a basic question open.
Does a language-trained RM use its state the way its mechanism intends?
Simply told what will matter, does it change what it saves to memory?
We test this with a task we call ``cued key--value recall'': a typical key--value recall task, but where the key is partially or completely revealed by a cue given before the list.
We find that models respond to exact matches but barely process the meaning of the cue when deciding what to save. Trained from scratch on a synthetic version of the task, the same architectures indeed behave rationally: they still save by exact match, but use the cue far more effectively than language-trained RMs, relying on it exactly when their state can no longer hold everything. Finally, limited memory does not only hurt.
After reading a hundred papers, you have forgotten almost every detail, yet you have likely built a far richer set of features and a compressed model of the field (Fig.~\ref{fig:overview}c).
Cognitive scientists have similarly argued that limits on processing can help a learner find structure \citep{hertwig2003more,cochran1999too}.
We find the same in neural networks: just as compression pressure during training is thought to drive representation learning, compression pressure at inference time pushes an RM toward better features and concepts. \textbf{Overall, our contributions are:}
\begin{enumerate}[leftmargin=1.5em,topsep=2pt,itemsep=1pt,parsep=0pt,partopsep=0pt]
    \item \textbf{Pretrained RMs mostly lack sophisticated memory control (Sec.~\ref{sec:real-models}, Fig.~\ref{fig:real-pretrained}, \ref{fig:real-finetuned}):} In a cued recall task, only an exact-match cue improves recall. Cues whose meaning must be semantically processed do nothing, and telling a model to ignore a record gains similarly to telling it to remember it. Fine-tuning strengthens both cued and uncued recall, but the exact match property mostly remains.
    \item \textbf{Synthetic experiments showing how different architectures modulate their memory (Sec.~\ref{sec:synthetic}, Fig.~\ref{fig:synthetic}):} We find that model architectures train quite rationally: RWKV-7 learns to rely on the cue when its state can no longer hold every record, and uncued recall is limited by memory capacity while cued recall is limited by length generalization. Mechanistically, we find that the cue is held in the first layer's state and used by the layers above to decide what to store.
    \item \textbf{Inference-time compression pressure can improve feature and concept learning (Sec.~\ref{sec:features}, Fig.~\ref{fig:mnist-basic}, \ref{fig:mnist-downstream}):} We show that a limited inference-time memory can force the model to learn more features. Trained only to recall the pixels it is given, an RM learns features that complete unseen digits and characters and support classification and concept readout, while a Transformer trained on the same data and loss simply copies.
    \item \textbf{A phase diagram of inference-time compression pressure (Sec.~\ref{sec:phase}, Fig.~\ref{fig:phase-diagram}):} We draw a phase diagram which shows how the inference-time compression pressure is clearly modulated by memory size and by how much the cue reveals about the query. The pressure helps in a middle regime: a moderate memory paired with an uninformative cue.
\end{enumerate}

\section{Preliminaries}
\label{sec:prelim}

We discuss related work in App.~\ref{sec:related}. One can write a fixed-state sequence model as:
\begin{equation}
s_t = F_\theta(s_{t-1},x_t), \qquad y_t=G_\theta(s_t,x_t),
\label{eq:recurrence}
\end{equation}
where the dimensionality of $s_t$ does not grow with $t$.
For a multi-layer model, $s_t$ collects the states of all layers and each layer updates its state from its previous state and the output of the layer below. RWKV carries per-layer matrix states; Mamba uses selective
state-space dynamics; Gated DeltaNet combines decay with a delta-rule update
\citep{gu2024mambalineartimesequencemodeling,
peng2025rwkv7gooseexpressivedynamic,
yang2025gateddeltanetworksimproving}. A full-attention Transformer can only be written in this form if the ``state''
contains the whole key--value cache, which grows linearly with the context.

It is worth pointing out that at least two layers are required to do \emph{state} (and thus arbitrary history) dependent saving in the linear RMs we study, since a layer's gates are computed from its input rather than from its state.
A lower layer can, however, pass a stored cue up to the gates of the layer above, which is how the trained model implements cued saving (More in Sec.~\ref{sec:synthetic}).
Nonlinear RNNs such as LSTMs compute their gates from the previous state and can do this within one layer.

\section{Recurrent Models mostly respond to exact match cues}
\label{sec:real-models}

\subsection{Cued key--value recall}
\label{sec:task}

\begin{figure}[t]
    \centering
    \includegraphics[width=0.9\textwidth]{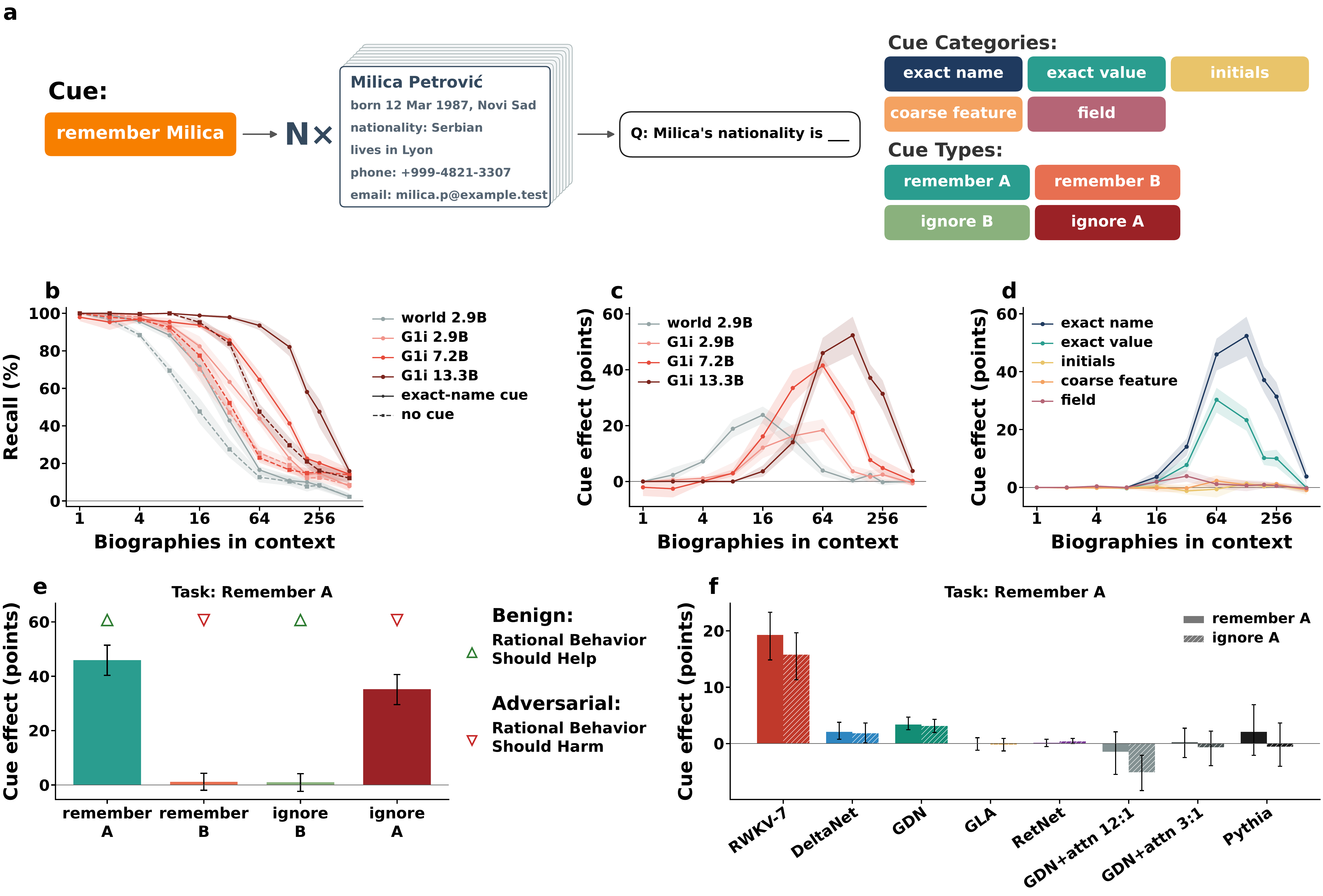}
    \caption{\textbf{Pretrained RMs mostly respond only to exact match cues.}
    \textbf{a.} The cued key--value recall task. Similar to a biography based key--value recall task, but a cue narrows down who the question will be about. The cue can be more or less precise, but could also be adversarial.
    (b) Recall accuracy with an exact-name cue and without for four pretrained RWKV-7 checkpoints (World 2.9B and G1i 2.9B, 7.2B, 13.3B) across different biography sizes.
    (c) The isolated cue effect grows with scale and peaks at longer contexts in larger models.
    (d) For RWKV G1i 13.3B, only cues that repeat exact text from the record help: the exact name or an exact field value. Initials, coarse features and the queried field have no effect.
    (e) When the task is to remember A, telling the model to ignore A still helps recall of A by 35 points, most of the 46-point benefit of telling it to remember A, where a rational policy would be harmed rather than helped. Cues about another person have no effect.
    (f) Remember-A and ignore-A cue effects on different models. Only RWKV-7 shows an effect, and negation barely reduces it. Shading and error bars: 95\% bootstrap intervals over eight content banks.}
    \label{fig:real-pretrained}
\end{figure}

We design a simple ``cued key--value recall'' task (Fig.~\ref{fig:real-pretrained}a) where the context is a list of $N$ short fictional biographies, followed by a question about one attribute of one person.
Each biography consists of a person's name, birth date and place, nationality, residence, phone number and email.
Unlike typical key--value recall \citep{graves2014neuralturingmachines,ba2016usingfastweightsattend} (also sometimes called needle in a haystack \citep{kamradt2023needle}), the model gets a cue \emph{before} the list arrives.
The cue narrows down which person the question will be about: it can give the person's exact name, an exact value from their profile, their initials, a coarse feature such as their birth decade, or the attribute that will be asked.
The cue can also tell the model to ignore a person rather than remember them.

We generally measure recall accuracy across different numbers of biographies.
We also often report the cue effect in points: recall with the cue minus recall without it, on the same list.
Note that we explore both benign cues, which point toward the person asked about (remember A, or ignore another person B), and adversarial cues, which point away (remember B, or ignore A); a rational memory policy gains from the first and loses from the second.
For further details, see App.~\ref{app:lm-eval}, \ref{app:bio-controls}.

\subsection{Pretrained RWKV-7 mostly responds to exact-match cues}

\emph{First of all, and perhaps most surprisingly, among the many language-pretrained models we test, only RWKV-7 reliably responds to any cue!}
At 1.3--1.5B parameters and 16 biographies, the cue effects for DeltaNet \citep{yang2025parallelizinglineartransformersdelta}, Gated DeltaNet \citep{yang2025gateddeltanetworksimproving}, GLA \citep{yang2024gatedlinearattentiontransformers}, RetNet \citep{sun2023retentivenetworksuccessortransformer}, two Gated DeltaNet--attention hybrids \citep{wang2026systematicanalysishybridlinear} and Pythia \citep{biderman2023pythiasuiteanalyzinglarge} are all within a few points of zero (Fig.~\ref{fig:real-pretrained}f; App.~\ref{app:roster}).
Note that this is not because the models are at ceiling: their uncued recall has already fallen to 5--10\%, and even closer to their own recall limit (1--8 biographies) the cue helps by at most about 13 points (Fig.~\ref{fig:app-roster}b, d).

For RWKV-7, the one model that listens, an exact-name cue starts helping when recall starts falling (Fig.~\ref{fig:real-pretrained}b, c).
As we scale the model, the cue effect starts later, since recall stays at ceiling longer, and the benefit grows: within the G1i line it peaks at 18 points for 2.9B and 42 for 7.2B (64 biographies), and at 52 for 13.3B (128 biographies).
However, we find that the model isn't really processing the cue semantically (Fig.~\ref{fig:real-pretrained}d, G1i 13.3B).
Only cues with an exact matching string elicit any cue effect: the full name, or an exact value from another field of the record (up to 30 points).
Any other cue, e.g., the person's initials, a coarse feature such as their birth decade, or the field that will be asked, does not help recall.
Furthermore, as seen in Fig.~\ref{fig:real-pretrained}e, negative cues, which should harm recall, in fact help almost as much as positive ones: telling the model to ignore A still raises recall of A by 35 points, against 46 for telling it to remember A.

\textbf{Overall, pretrained RMs mostly ignore the cue, and RWKV-7, the one that uses it, does so by exact matching rather than by understanding it.}

\subsection{Fine-tuned RMs follow the cue, but still mostly by matching its tokens}

\begin{figure}[t]
    \centering
    \includegraphics[width=\textwidth]{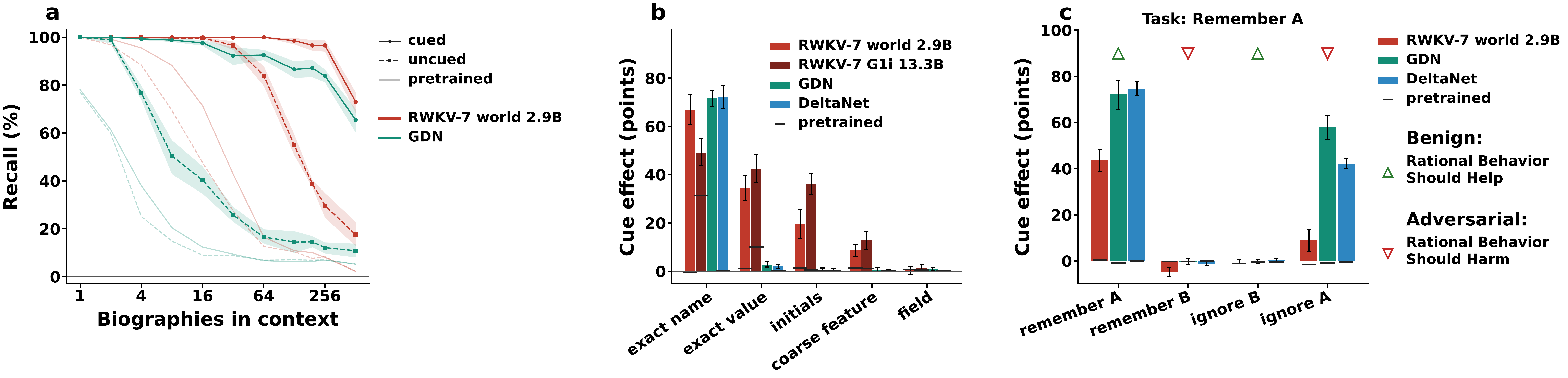}
    \caption{\textbf{Fine-tuning makes the cue work, but how much of its meaning is used depends on the architecture.}
    Models are fine-tuned on benign cues only, including vague ones (initials, coarse features, the queried field) and negative ones (``ignore B''); faint lines and black ticks show the same models before fine-tuning.
    \textbf{a.} With an exact-name cue, RWKV-7 greatly improves both uncued and cued recall, while GDN improves mostly on cued recall.
    \textbf{b.} Cue effects by cue type at 256 biographies. GDN and DeltaNet still use only the exact name, whereas RWKV-7 also uses exact values, initials and, more weakly, coarse features. No model uses the field cue.
    \textbf{c.} Cue effects at 128 biographies when the question is about A. ``Remember A'' helps every model, but so does the adversarial ``ignore A'' (strongly for GDN and DeltaNet), while the benign ``ignore B'' does nothing, despite explicit training on such negative cues.
    Shading and error bars: 95\% bootstrap intervals over eight content banks.}
    \label{fig:real-finetuned}
\end{figure}

Disappointed by the models' response to the cue, we next try to fine-tune them for this behavior.
We fine-tune each model on 12{,}000 biography episodes, all with benign cues and correct answers (one training seed per model).
We find that for RWKV-7, fine-tuning greatly improves both uncued and cued recall: at 64 biographies uncued recall rises from 13\% to 84\%, and cued recall stays at 97\% even at 256 biographies, up from 8\% before fine-tuning, an 89-point gain (Fig.~\ref{fig:real-finetuned}a).
Gated DeltaNet instead improves mostly when cued: its uncued recall stays at 12--17\%, but cued recall reaches 84--93\% up to 256 biographies.
GLA and RetNet do not acquire such behavior (App.~\ref{app:bio-ft}).

However, semantic processing of the cue still \emph{mostly} remains disappointing (Fig.~\ref{fig:real-finetuned}b, c).
All models still show a positive effect of an adversarial cue, although smaller than that of the benign one: telling a model to ignore A still raises recall of A by 58 (Gated DeltaNet), 42 (DeltaNet) and 9 (RWKV-7) points (Fig.~\ref{fig:real-finetuned}c).
Most gains are also still limited to exact matches.
\emph{There is some hope, however: RWKV-7 does start responding to vague cues} such as initials (20 points for World 2.9B, 36 for G1i 13.3B) and coarse features (9 and 13), while Gated DeltaNet is strictly limited to the exact name (Fig.~\ref{fig:real-finetuned}b).
No model learns to use a cue naming the field that will be asked, even though such cues made up a sixth of the training data.
Note that we do not measure whether the fine-tuned models change what they write to state, or implement the cue through another query-time strategy.

\textbf{Overall, fine-tuning shows that the capacity to follow the cue was there all along, but what is learned stays largely token-based.}
Whether a recurrent architecture discovers the policy when trained from scratch, and which details of the architecture decide that, is what we test next in synthetic episodes where every factor can be controlled.

\subsection{Trained from scratch, RWKV-7 learns a rational cue policy}
\label{sec:synthetic}

\begin{figure}[t]
    \centering
    \includegraphics[width=0.9\textwidth]{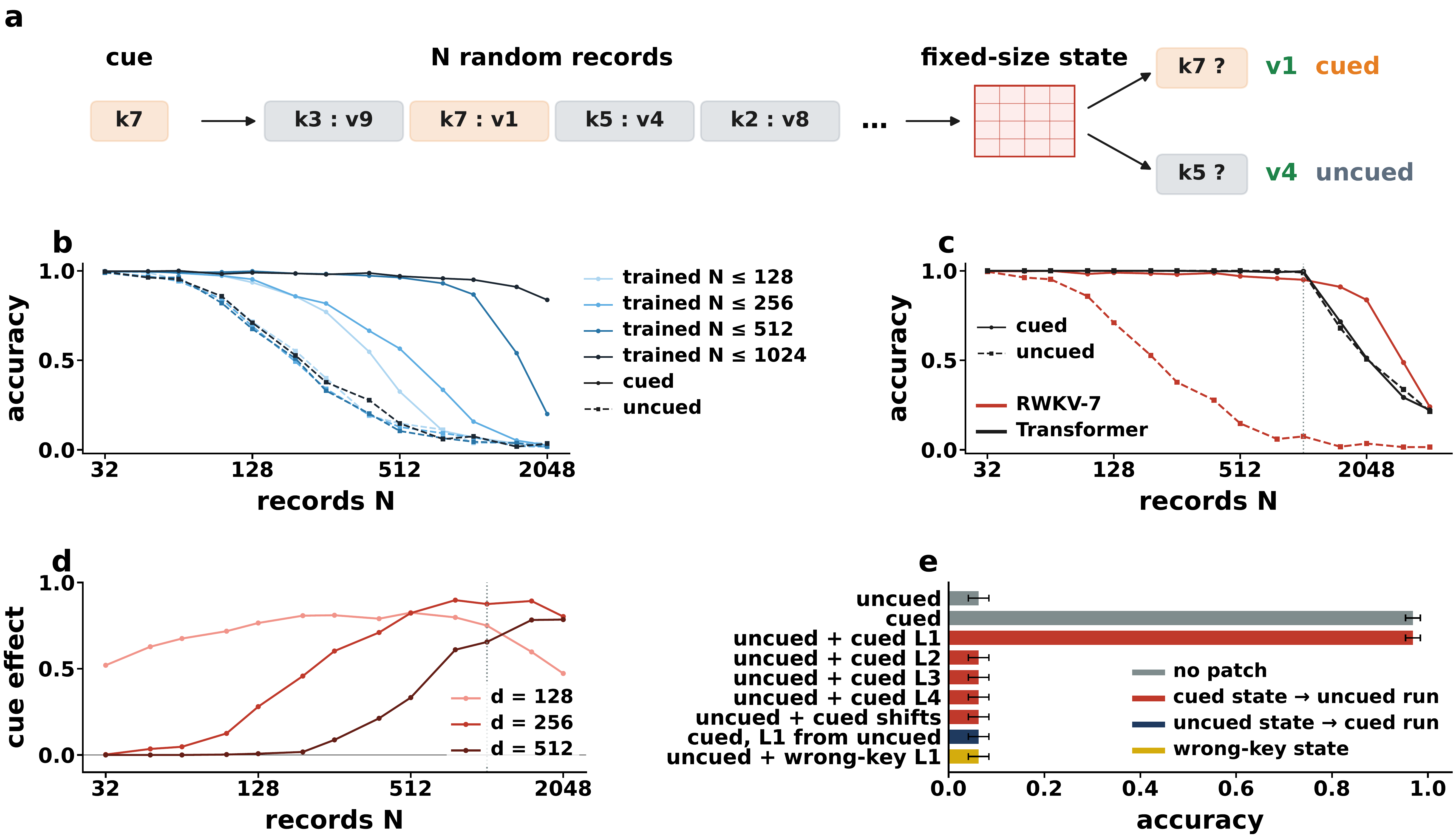}
    \caption{\textbf{Trained from scratch, RWKV-7 uses the cue rationally.}
    \textbf{a.} Synthetic cued key--value recall: a cue names one key, then $N$ random key--value records follow, and the query asks for the value of either the cued key or another key.
    \textbf{b.} RWKV-7 trained up to different lengths $N$. Uncued recall collapses onto one curve for every training length, set by memory capacity, while cued recall extends with the training length, set by length generalization.
    \textbf{c.} A Transformer recalls everything with or without the cue within its training length ($N\le1024$), so the cue buys it nothing, while RWKV-7 needs the cue once the records exceed its state (dotted line: training length, $N=1024$).
    \textbf{d.} The smaller the state (hidden size $d$), the sooner the cue starts to matter.
    \textbf{e.} Transplanting just the layer-1 state after the cue into an uncued run's state restores cued recall, while layers 2--4, token shifts or a wrong-key state do not. Also, a cued run with an uncued layer-1 state fails to recall.
    Error bars: standard error over 128 documents. One training seed per model.}
    \label{fig:synthetic}
\end{figure}

Since RWKV-7 was the most promising model so far, we now ask whether it can learn the cue policy from scratch, in a synthetic version of the task where nothing but the cue tells the model what to keep.
Each episode is a cue key, then $N$ random key--value records, then a query (Fig.~\ref{fig:synthetic}a); since keys and values are random tokens, no record is more salient than another.

And indeed, RWKV-7 learns it.
A small four-layer RWKV-7 (16{,}384 state floats) trained up to $N=1024$ recalls 95\% of cued records at $N=1024$ but only 7.5\% of uncued ones (Fig.~\ref{fig:synthetic}c).
A Transformer, on the other hand, recalls everything with or without the cue within its training length ($N\le1024$), so the cue buys it nothing (App.~\ref{app:formula-sweep}).
Training on longer lists, uncued recall always collapses along the same curve, set by memory capacity, while cued recall extends with the training length, set by length generalization (Fig.~\ref{fig:synthetic}b).

The policy is also rational: the cue starts to matter once the state is full, so the smaller the state, the sooner this happens (Fig.~\ref{fig:synthetic}d).
At $N=128$, the cue is worth 0.77 for hidden size $d=128$, 0.28 for $d=256$ and almost nothing for $d=512$, whose state still holds most records.
Interestingly, a model whose state holds its whole training range may never learn the cue at all: $d=512$ trained up to $N=512$ shows a cue effect of at most 0.04, but reaches 0.79 when trained up to $N=1024$ (App.~\ref{app:width}; one seed per cell).

Finally, the cue lives in the layer-1 state (Fig.~\ref{fig:synthetic}e).
Transplanting only the layer-1 state after the cue into an uncued run restores cued recall, while layers 2--4, token shifts or a wrong-key state do nothing.
Layer 1 holds the cue, and the layers above use it to decide what to store (App.~\ref{app:causal}).
Other RMs are less reliable: with a matched state, Gated DeltaNet learns the policy only when warm-started from shorter training or trained three times longer, and DeltaNet in two of three seeds (App.~\ref{app:arch-budget}).

\textbf{Overall, trained on data that rewards it, RWKV-7 saves what the cue names exactly when its state can no longer hold everything.}

\section{Inference-time pressure can improve learned features}
\label{sec:features}

\subsection{From known relevance to unknown future use}
Cued recall assumes that the future relevance is partly given during inference. But what should a bounded model do when the future task is not known? We argue the best it can do is to learn the shared structure underying the task to store the incoming evidence efficiently.

This is directly related to the common statement ``intelligence is compression''. This statement is often framed as a pressure during pretraining, but here we argue that an \emph{inference-time compression pressure}, exists for recurrent models.

We therefore ask the following: \emph{can inference-time pressure change the learned solution for the same task/learning rule?}

\subsection{Inference-time pressure forces a generalizable summary state}
\label{sec:mnist}

\begin{figure}[t]
    \centering
    \includegraphics[width=0.9\textwidth]{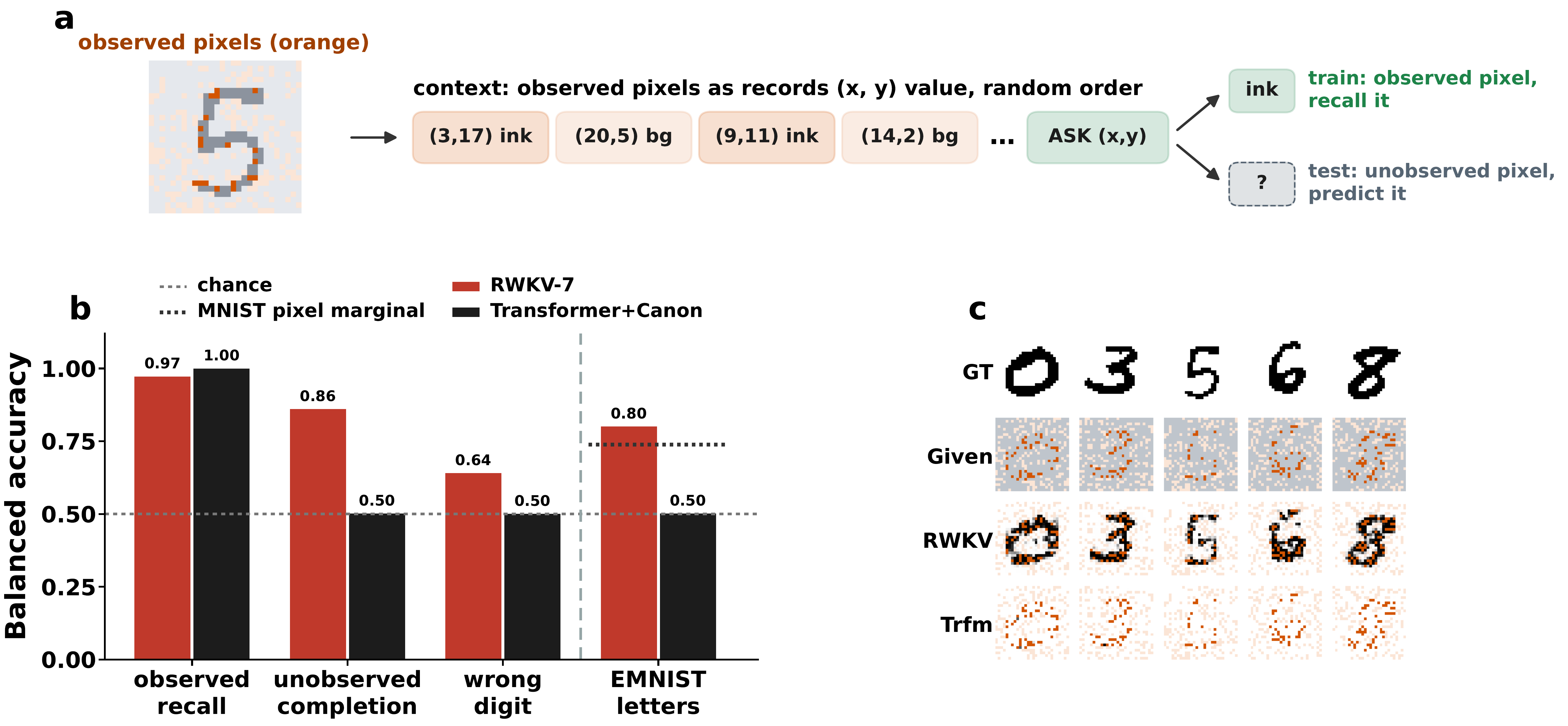}
    \caption{\textbf{Inference-time pressure forces models to learn generalizable summary states.}
    \textbf{a.} Partial pixels of an MNIST digit are given as $(x, y, value)$ records in random order. During training \emph{we only query for observed pixels} so copying solves it, if the model has enough memory.
    \textbf{b.} Both the Transformer and RWKV-7 manage to recall seen pixels, but only RWKV-7 completes unobserved ones, and this also transfers to EMNIST letters. The Transformer stays at chance level.
    \textbf{c.} RWKV-7 can roughly reconstruct missing pixels from random conditionings while the Transformer returns only the given pixels; more examples in App.~\ref{app:gallery}.}
    \label{fig:mnist-basic}
\end{figure}

We devise a task (Fig.~\ref{fig:mnist-basic}a) where \emph{random pixels} from a binarized MNIST \citep{Lecun_1998} digit are given as a list of $(x, y, \mathrm{value})$ tuples in random order.
During training, we only ask for pixels that are already in the context.
Thus literal key--value recall is a perfect solution, since missing pixels never contribute to the loss.

We then test the model on pixels which are \emph{not} specified by the context.
First, quantitatively, we find that both models recall observed pixels (balanced accuracy 0.97 for RWKV-7 and 1.00 for a parameter-matched Transformer with Canon layers), but only RWKV-7 completes the missing ones: 0.86, against 0.50 for the Transformer, which simply predicts background (Fig.~\ref{fig:mnist-basic}b).
This completion uses the episode itself: replacing the context with an ink-matched image of a different digit drops RWKV-7 to 0.64.
Qualitatively, the reconstructions in Fig.~\ref{fig:mnist-basic}c clearly show that the Transformer barely predicts ink outside of what the context specifies, while RWKV-7 completes the digit.
The MNIST-only state even transfers to EMNIST \citep{Cohen_2017} letters, reaching 0.80 on missing pixels against 0.74 for the MNIST pixel-frequency baseline and 0.50 for the Transformer (Fig.~\ref{fig:mnist-basic}b; App.~\ref{app:emnist}); this is evidence for reusable spatial structure, not letter recognition.
All comparisons here use one training seed per architecture.

\textbf{Overall, trained only to recall what it has seen, RWKV-7 learns a summary of the digit that completes what it has not seen, while a Transformer that can simply copy learns nothing beyond the copy.}

\subsection{Inference-time pressure also yields features and concepts}
We now ask a deeper question: does inference-time pressure merely force a general smoothing of the data, or does it learn actual features and concepts?

\begin{figure}[t]
    \centering
    \includegraphics[width=\textwidth]{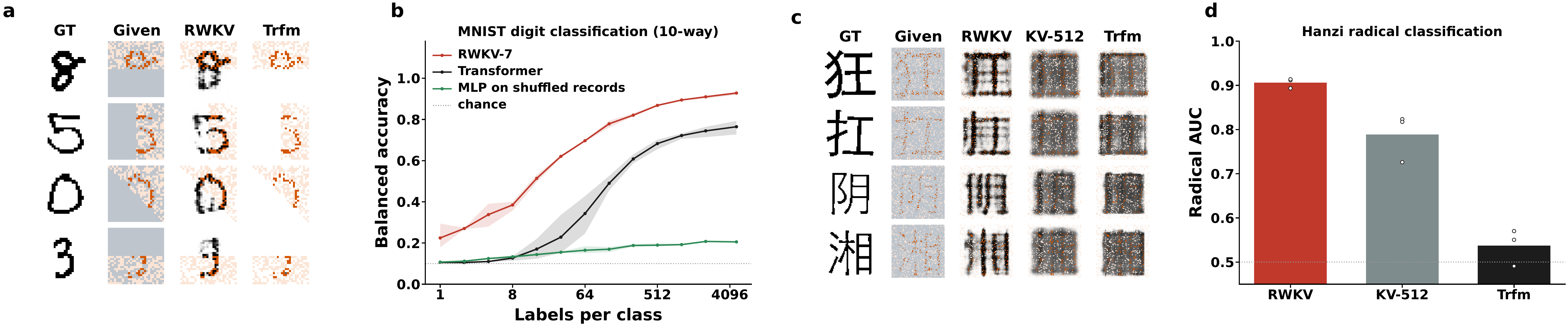}
    \caption{\textbf{Features learned under inference-time pressure support completion, classification and concepts.}
    \textbf{a.} With large regions hidden, RWKV-7 draws plausible missing strokes, while the Transformer returns only the given pixels.
    \textbf{b.} A classification token trained on top of the recall-trained RWKV-7 needs far fewer labels to classify digits than one on the Transformer, or an MLP given the same data format.
    \textbf{c.} On held-out Hanzi characters, RWKV-7 reconstructs stroke structure, while Transformers with bounded (KV-512) or full attention do not in these examples (seed 0); on hidden-pixel accuracy the bounded Transformer matches RWKV-7 in two of three seeds (App.~\ref{app:hanzi}).
    \textbf{d.} A Hanzi character's radical is linearly readable from the RWKV-7 state, less so for KV-512 and barely for full attention (dots: three seeds; chance 0.5).}
    \label{fig:mnist-downstream}
\end{figure}

To test this, we hide large parts of MNIST digits, building contexts unlike any seen during training (Fig.~\ref{fig:mnist-downstream}a).
\emph{Surprisingly, RWKV-7 completes digits far outside the given region!}
It draws the whole lower loop of the 8, and completes the 3 whose top half is hidden as a 9, a reasonable guess from what is visible, while the Transformer again returns only the given pixels.

The learned state also makes classification more label-efficient.
We train a new classification token and head on top of the recall-trained models (Fig.~\ref{fig:mnist-downstream}b): with 16 labels per class, RWKV-7 reaches 0.51 balanced accuracy against 0.17 for the Transformer (0.93 against 0.76 with all labels), while an MLP trained on the same shuffled records barely learns.
This advantage is relative to models that read serialized records; a masked-recall autoencoder that sees the 2-D image does better (App.~\ref{app:input-matched}).

Finally, we test on Hanzi, a script with far richer structure.
Trained only to recall observed pixels of 930 characters, RWKV-7 reconstructs the stroke structure of held-out characters, while a Transformer with full memory does not, and one with a bounded memory (512 key--value tokens) does so only in some seeds (Fig.~\ref{fig:mnist-downstream}c; App.~\ref{app:hanzi}).
A linear probe on the RWKV-7 end-of-context state also reads out the characters' radicals with macro AUC 0.91 (mean of three seeds), against 0.79 for the bounded Transformer and 0.54 for full attention (Fig.~\ref{fig:mnist-downstream}d; App.~\ref{app:hanzi}).
The raw visible pixels give 0.95, so the state keeps the radical information in the context rather than adding to it; for full attention, the end-of-context state is also not where its memory lives.

\textbf{Overall, the state learned under inference-time pressure is more than a smoothing: it completes digits from their shape, makes classification label-efficient, and keeps concept-level information such as Hanzi radicals.}

\subsection{A phase diagram of inference-time pressure}
\label{sec:phase}

Finally, we combine the two threads of this paper, informative cues about the future and inference-time pressure, and show that how much pressure a model feels depends strongly on both the architecture and the task.

\begin{figure}[t]
    \centering
    \includegraphics[width=0.95\textwidth]{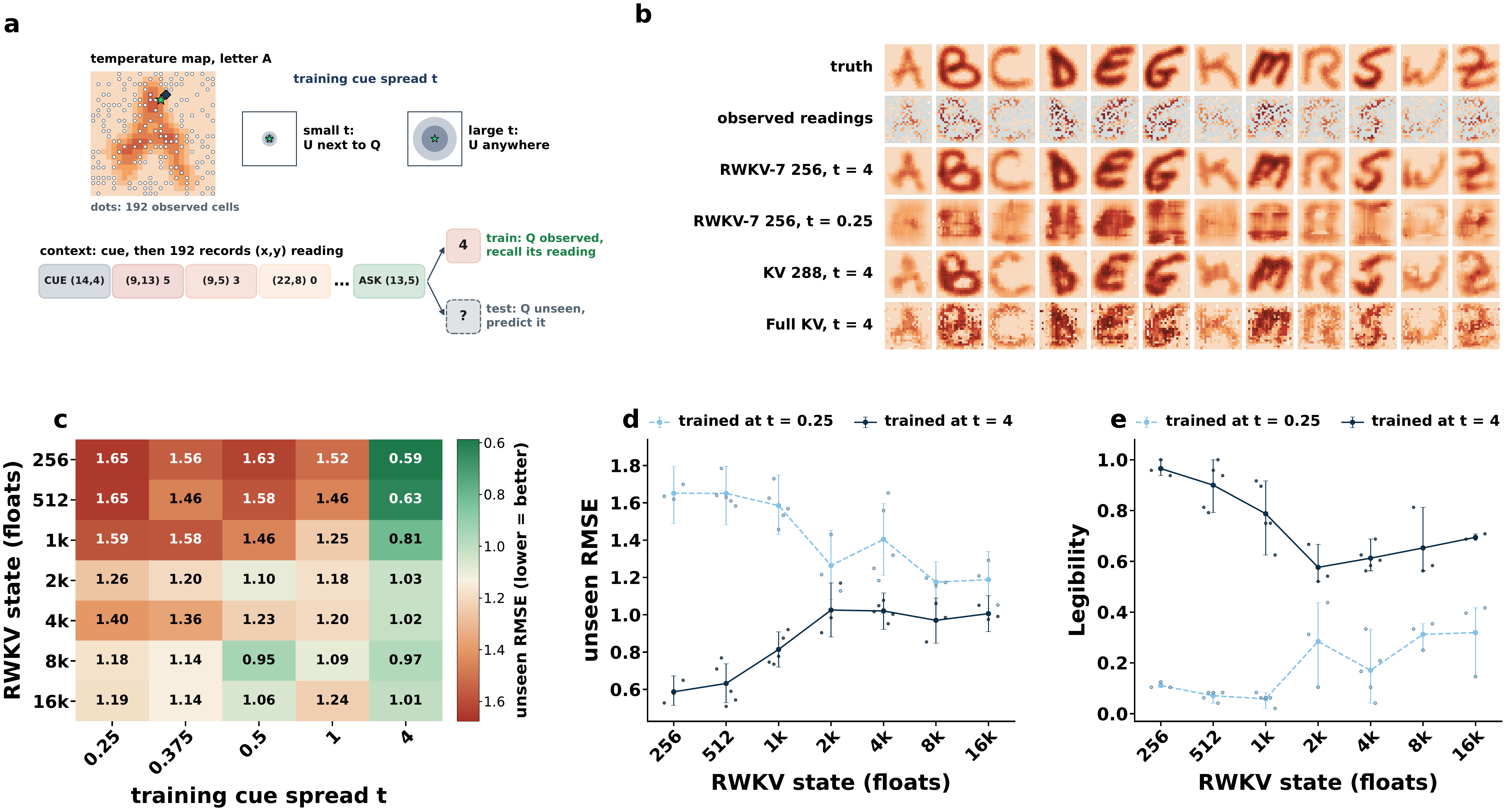}
    \caption{\textbf{A phase diagram of inference-time pressure.}
    \textbf{a.} A letter shaped heat map is observed at $\simeq1/3$ pixels and a cue precedes the list of evidence given as ($x,y,value$). The cue specifies the to-be-queried pixel with an error of $t$ and sets how much it reveals about the query $Q$. Thus small $t$ is very informative while big $t$ isn't so much informative.
    \textbf{b.} A small-state RWKV-7 trained with an uninformative($t=4$) cue and a Transformer with a 288-token KV-budget reconstruct the letters; while the same RWKV-7 trained with more informative cues($t=0.25$) and a Transformer with full attention, do not do so as well.
    \textbf{c--e.} The phase diagram of inference-time pressure in action. For RWKV-7, a small state helps only when the training cue spoils little about the query. Same experiments for Transformers with a bounded key--value budget are in App.~\ref{app:phase-kv}.
    Dots: training seeds; error bars: 95\% bootstrap intervals.}
    \label{fig:phase-diagram}
\end{figure}

To this end, we design a task similar to the MNIST one (Fig.~\ref{fig:phase-diagram}a).
A letter-shaped heat map on a $24\times24$ grid is observed at 192 random pixels, given as a list of $(x, y, \mathrm{value})$ readings, and a cue before the list points to the pixel that will be queried, with a spread $t$.
As in Sec.~\ref{sec:mnist}, during training we only ask the model to regurgitate observed pixels, and at test time we query pixels not given in the context.
The amount of memory necessary for the task is now controlled by two variables: how much the model knows about the future query, and how much memory capacity it has.
We therefore expect more pressure as $t$ gets larger and as the state gets smaller.

Qualitatively, as expected, clean letters form when the cue is uninformative and the state is small (Fig.~\ref{fig:phase-diagram}b).
The phase diagram quantifies this (Fig.~\ref{fig:phase-diagram}c, d): RWKV-7 trained with an uninformative cue reconstructs unseen pixels best with the smallest states (RMSE 0.59 at 256 floats, 0.63 at 512), the only settings that beat interpolating the four nearest observed readings (0.65), while the same small states trained with an informative cue ($t=0.25$) fail (1.65).
For completeness, we also measure how identifiable the letters are with a frozen letter classifier: it reads 97\% of the 256-float, $t=4$ reconstructions correctly, against 11\% at $t=0.25$ (Fig.~\ref{fig:phase-diagram}e).

Interestingly, a Transformer given an artificial inference-time pressure, a bounded key--value budget, also learns better features: with 288 tokens it reconstructs the letters (RMSE 0.73, 85\% legible), while full attention does not (1.63, 38\%), though the cue matters much less for it (Fig.~\ref{fig:phase-diagram}b; App.~\ref{app:phase-kv}).

\textbf{Overall, inference-time compression pressure, whether from an uninformative cue or a small memory, can push a model toward features and concepts that generalize, when the data have structure to compress; ours is a controlled existence result, not a universal compression principle.}

\section{Discussion}
\label{sec:discussion}

\paragraph{Why does language pretraining not teach cued saving?}
It is intriguing that a perfectly capable architecture (Sec.~\ref{sec:synthetic}) does not learn cue-based memory saving from pretraining, a behavior any human easily demonstrates.
We think that natural-language pretraining data rarely highlights such an ability: text seldom announces in advance what will be asked about later, and even more rarely rewards saving something because of such an announcement, let alone understanding a semantic cue or a negation.
When text does refer back, it usually repeats the exact name or phrase, which is exactly the token-matching policy we observe.
Perhaps more fundamentally, text mostly follows the arrow of time: next-token prediction rewards predicting what comes next from what came before, and a later question is rarely \emph{caused} by an earlier instruction to remember.
For cued saving, pretraining data is in this sense simply boring, and our fine-tuning results (Sec.~\ref{sec:real-models}) suggest that a modest amount of data that does reward it is enough to elicit it.

\paragraph{Why can inference-time memory pressure be an advantage?}
A limited memory, trivially, loses detail as more information arrives, which clearly seems like a limitation. However, we argue that limited memory is not always a disadvantage, especially when retrieval or storage tools are available \citep{malach2025infinitybeyondtooluseunlocks}. The most intuitive example is a scholar who has just read 100 papers. They probably cannot recall a single abstract perfectly, but they are most likely equipped with a rich, compressed model of the field. We think this is the right way to view inference-time pressure: as compressing a stream into a consistent world model \citep{schmidhuber2009drivencompressionprogresssimple}.

\paragraph{Limitations.}
\label{sec:limitations}
Our work is an analysis: we do not show real-world performance gain, our from-scratch experiments (Secs.~\ref{sec:synthetic} and~\ref{sec:features}) use small models and synthetic tasks as proofs of concept, and for language models we mostly measure recall rather than state writes.
Despite our experiments across architectures and recipes, it is still not completely clear what determines whether a model can learn cued saving.
Finally, our state sizes and key--value budgets are operational limits rather than formal bit budgets.

\clearpage
\subsection*{AI use statement}
We used generative AI tools for:
\begin{itemize}
    \item \textbf{Code and analysis:} writing experimental, analysis, and plotting code. All code
    was checked by the authors and results were verified with control experiments; most were
    reproduced independently by multiple independent authors.
    \item \textbf{Experiments:} running variations around experiments specified by the
    authors. The hypotheses, experimental designs, and interpretations are the authors' own.
    \item \textbf{Writing:} polishing text, suggesting phrasings, and finding literature. The
    text, claims, and narrative are the authors' own.
\end{itemize}

Thus, we used AI for: \textit{implement methods}, \textit{create or edit software code}, \textit{create or modify scientific figures or images},
\textit{edit a research paper to improve readability},
and \textit{identify relevant literature}. We did not use them to \textit{propose or refine
hypotheses} or \textit{interpret results}. We have reviewed all AI-assisted work and take
responsibility for the final content of this work, including text, claims, and artifacts produced
with the aid of generative AI.

\subsection*{Ethics statement}

This paper presents work whose goal is to advance the field of Machine Learning. There are
many potential societal consequences of our work, none of which we feel must be specifically
highlighted here.

\subsection*{Reproducibility statement}
The cued key--value recall task is described in Sec.~\ref{sec:task}, with full language-model evaluation and fine-tuning details in App.~\ref{app:lm-eval}. The synthetic key--value recall setup is in App.~\ref{app:synthetic-setup}, and the MNIST/EMNIST pixel recall, Hanzi, sliding-window control and letter-heat phase-diagram setups are in App.~\ref{app:pixel-setup}. All code will be released.

\bibliography{iclr2027_conference}

\appendix

\noindent Table~\ref{tab:appmap} summarizes the contents of each appendix and the main-text material it supports.

\begin{table}[h]
\centering
\caption{\textbf{Appendix guide.}}
\label{tab:appmap}
\footnotesize\begin{tabular}{lll}
\toprule
Appendix & Contents & Supports \\
\midrule
\ref{sec:related} & Related work & All sections \\
\midrule
\ref{app:lm-eval} & LLM biography recall: experimental details & Sec.~\ref{sec:real-models} \\
\ref{app:roster} & LLM biography recall: additional results & Figs.~\ref{fig:real-pretrained}, \ref{fig:real-finetuned} \\
\midrule
\ref{app:synthetic-setup} & Synthetic key--value recall: experimental details & Sec.~\ref{sec:synthetic} \\
\ref{app:formula-sweep} & Synthetic key--value recall: additional results & Fig.~\ref{fig:synthetic} \\
\midrule
\ref{app:pixel-setup} & Pixel recall and compression: experimental details & Secs.~\ref{sec:features}--\ref{sec:phase} \\
\ref{app:gallery} & Pixel recall and compression: additional results & Figs.~\ref{fig:mnist-basic}--\ref{fig:phase-diagram} \\
\bottomrule
\end{tabular}
\end{table}

\clearpage
\section{Related work}
\label{sec:related}

\textbf{Efficient recurrent sequence models.} RWKV, Mamba, Mamba-2 and Gated DeltaNet keep a fixed-size state instead of a KV-cache that grows linearly with context \citep{peng2025rwkv7gooseexpressivedynamic, gu2024mambalineartimesequencemodeling, dao2024transformersssmsgeneralizedmodels, yang2025gateddeltanetworksimproving}. The computations these models can express depend on the state transition rule, more specifically on the eigenvalues of the transitions \citep{grazzi2025unlockingstatetrackinglinearrnns}. In this work, instead of asking what's possible, we ask whether these memory modules behave intuitively in the way they were designed for, and what data and training are required for them to do so.

\textbf{Linear attention, state tracking, copying and tools.} Non-quadratic, $O(1)$ inference-cost language models have always been of interest due to inference efficiency, as well as motivations in continual learning \citep{schlag2021lineartransformerssecretlyfast, irie2023metalearningcontinuallearning} and length generalization. Linear attention allows parallel training and constant-cost inference by removing the softmax in the self-attention computation \citep{katharopoulos2020transformersrnnsfastautoregressive}:

\begin{equation*}
S_t = S_{t-1} + \phi(k_t) v_t^\top, \qquad z_t = z_{t-1} + \phi(k_t), \qquad y_t = \frac{\phi(q_t)^\top S_t}{\phi(q_t)^\top z_t}.
\end{equation*}

Mamba focused on enabling input-dependent state updates \citep{gu2024mambalineartimesequencemodeling}, and the delta rule allowed states to overwrite old associations with new ones \citep{yang2025parallelizinglineartransformersdelta}. However, interestingly, these models, despite their naming ``state space models'', were not able to track state \citep{merrill2025illusionstatestatespacemodels} until \citet{grazzi2025unlockingstatetrackinglinearrnns} identified that a single-line change allowing negative eigenvalues fixes this. Closer to the motivation of our paper, it has also been debated how much this finite state matters. Initially, researchers identified that (not surprisingly) RMs are worse than Transformers at pure copying \citep{jelassi2024repeatmetransformersbetter}, however later work argued that this might not be such an issue when provided with external tools \citep{malach2025infinitybeyondtooluseunlocks}, just like people rely on external notes and compute.

\textbf{Associative recall and long context.} Associative-recall benchmarks have shown that efficient sequence models are often suboptimal at retrieval tasks \citep{arora2023zoologymeasuringimprovingrecall}, and interestingly, the same task can be implemented by quite different mechanisms depending on the architecture \citep{arora2026mechanisticevaluationtransformersstate}. Closest to our setting, Just Read Twice \citep{arora2024justreadtwiceclosing} shows that for a recurrent LM, the order of context and question matters for deciding what to store: simply repeating the prompt lets the model read it again knowing the question, which improves recall. Transformers are not free of such issues either. Their long-context recall depends on where the relevant passage sits \citep{liu2023lostmiddlelanguagemodels}, and drops sharply when the question shares no literal words with the passage \citep{modarressi2025nolimalongcontextevaluationliteral}, not unlike the token-matching behavior we observe. Furthermore, larger LMs also struggle with negated instructions \citep{jang2022largelanguagemodelstruly}.

\textbf{In-context algorithms and test-time learning.} Whether an in-context learner retrieves or infers has been shown to shift with task diversity, context length and training time \citep{park2025competitiondynamicsshapealgorithmic, wurgaft2025incontextlearningstrategiesemerge}. In Transformers, in-context learning can even emerge and then vanish while the training loss keeps falling \citep{singh2023transientnatureemergentincontext}. In hybrids, a short attention window during training forces the recurrent layers to learn long-term memory \citep{cabannes2026shortwindowattentionenables}. Test-time learning takes this one step further and trains the state as a memory while reading \citep{sun2025learninglearntesttime, behrouz2024titanslearningmemorizetest}. However, standard memory evaluations such as needle in a haystack \citep{kamradt2023needle} ask for the needle only after the content, so they cannot tell whether a memory writes what an advance cue asks for. In this work, we vary the finite inference state as another knob, and ask whether it pushes a model to copy literally, to store only what a cue asks for, or to build a compressed representation of the episode.

\textbf{Memory constraints and cognition.} Our cognitive analogy is functional, not mechanistic: current RMs are not biological models, and a limit in cognition does not automatically imply a benefit in neural networks. That said, the idea that limits can help has a long history in cognitive science. \citet{newport1990maturational} proposed that children learn language better partly \emph{because} of their limited cognitive capacity, \citet{cochran1999too} argued that a larger processing capacity can be a disadvantage for adult learners, and \citet{hertwig2003more} argued more broadly for the benefits of cognitive limits. In neural networks, \citet{elman1993learning} found that a simple RNN learned a complex grammar only when its working memory started limited, whereas \citet{rohde1999language} found that starting small was unnecessary, and even hindered learning, on more English-like grammars. Thus, we do not assume that limits help; instead, we test when a bounded inference state does.

\clearpage
\section{LLM biography recall task}
\label{app:exp-real}
\label{app:setup}

This section provides experimental details and additional results for the language-model biography recall experiments in Sec.~\ref{sec:real-models} (Figs.~\ref{fig:real-pretrained} and~\ref{fig:real-finetuned}).

\subsection{Experimental details}
\label{app:lm-eval}
\label{app:notation}
\label{app:lm-ft}

\paragraph{Models.}
We evaluate 15 pretrained language models:
\begin{itemize}
    \item RWKV-7 g1 0.4B, 1.5B, and 2.9B \citep{peng2025rwkv7gooseexpressivedynamic};
    \item RWKV-7 G1i 2.9B, 7.2B, and 13.3B \citep{peng2025rwkv7gooseexpressivedynamic};
    \item RWKV-7 World 2.9B \citep{peng2025rwkv7gooseexpressivedynamic};
    \item DeltaNet \citep{yang2025parallelizinglineartransformersdelta}, Gated DeltaNet \citep{yang2025gateddeltanetworksimproving}, GLA \citep{yang2024gatedlinearattentiontransformers}, and RetNet \citep{sun2023retentivenetworksuccessortransformer} at 1.3B, trained on 100B tokens;
    \item Two Gated DeltaNet--attention hybrids (12:1 and 3:1 linear-to-attention layers) from the same suite \citep{wang2026systematicanalysishybridlinear};
    \item Pythia 410M and 1.4B \citep{biderman2023pythiasuiteanalyzinglarge}.
\end{itemize}
RWKV-7 and the 1.3B models are run in bfloat16 precision, while Pythia runs in float32. We use the official chat template for RWKV-7: RWKV-7 World uses \texttt{User: \dots\textbackslash n\textbackslash nAssistant:}, and g1/G1i use the same template with an empty thinking block. Models without a chat template are evaluated with a common \texttt{User:/Assistant:} template. Note that Pythia and the two hybrid models have a native sequence length of 2{,}048 tokens; since 16 biographies take around 1{,}300 tokens and 32 biographies take around 2{,}500 tokens, their results beyond 16 biographies measure length generalization.

\paragraph{Biographies.}
We generate 1{,}024 fictional biographical profiles with Gemini~3.1 Flash-Lite \citep{geminiteam2024geminifamilyhighlycapable}. We take a few precautions so chance accuracy does not reflect natural priors: we balance the occurrences of sex, age, and nationality; for phone numbers we use the unassigned \texttt{+999} prefix; and for emails we use the reserved \texttt{.test} domain.

\begin{figure}[!htbp]
    \centering
    \includegraphics[width=\textwidth]{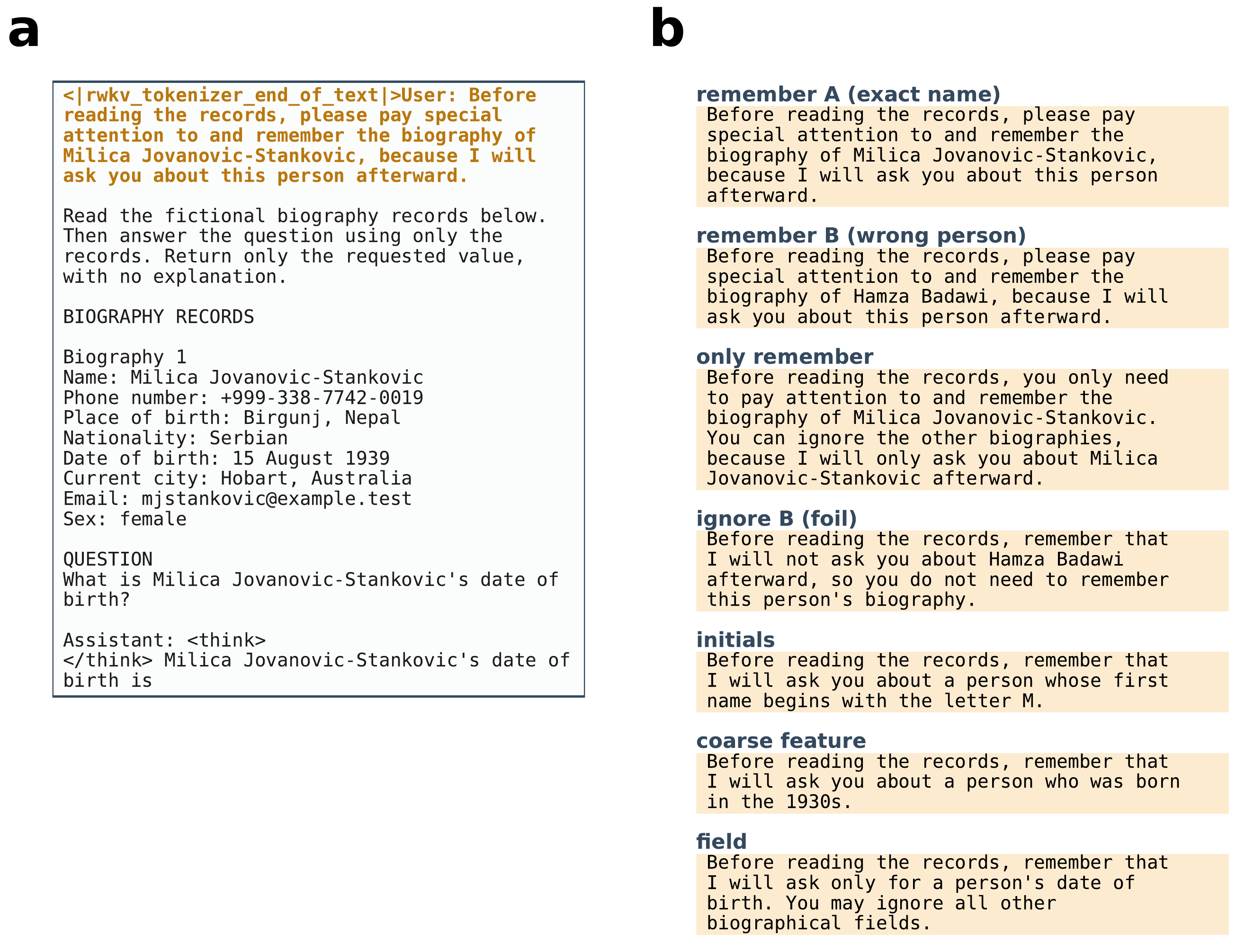}
    \caption{\textbf{The biography prompt.}
    \textbf{a.} A complete prompt with an exact-name cue and one biography; the cue sentence is highlighted.
    \textbf{b.} The cue sentence of each condition. Everything else in the prompt is identical across conditions.}
    \label{fig:app-prompt}
\end{figure}

Each biography has eight labeled fields (Fig.~\ref{fig:app-prompt}a) and we query six attributes: date of birth, nationality, place of birth, current city, phone number, and email. Each document we test contains $N\in\{1,2,4,\dots,256,512\}$ biographies, evaluated across eight independent sets of biographies. For each set, we evaluate four different orderings (two biography orders crossed with two field orders). Overall, each document has up to four target people and each target is asked about all six attributes, giving 352 documents and 7{,}488 questions per cue condition.

\paragraph{Cues.}
A single cue sentence is placed before the biographies (Fig.~\ref{fig:app-prompt}b). We test five types of cues:
\begin{enumerate}
    \item Full name;
    \item Initials: first initial, last initial, or both;
    \item One exact attribute value from the six queried fields, excluding the queried attribute itself;
    \item A coarse hint: birth decade, birth continent, phone prefix, or email format;
    \item The queried field.
\end{enumerate}
Among these, (1) and (3) are exact text matches from the biography, while (2), (4), and (5) give only partial or semantic information.

Assume A is the person the question will be about and B is another person in the list. Each cue type is evaluated under positive (e.g., ``I will ask about'') and negative (e.g., ``I will not ask about'') wording, and each cue can be either truthful or adversarial, giving 61 cue conditions in total. ``Remember A'' and ``ignore B'' are benign cues since they give helpful information about the query, so a rational memory policy should show a performance gain, whereas ``remember B'' and ``ignore A'' are adversarial and should hurt performance.

\paragraph{Scoring, cue effect, and confidence intervals.}
We score recall by greedy exact match on the model's generated answer, averaged over the six queried attributes. After the biographies and the question in the user turn (Fig.~\ref{fig:app-prompt}a), we prefill the start of the assistant reply with ``$\langle$name$\rangle$'s $\langle$attribute$\rangle$ is'' so the model only needs to output the value. Each document is encoded into a state once and we generate each answer from a copy of that state or cache. We stop generation at the end-of-text token, a line break or a role marker. The recalled attribute is then scored by exact match up to Unicode NFKC normalization, casing, whitespace and trailing punctuation.

We also check two alternative scoring controls in App.~\ref{app:bio-controls}: free-answer recall (Fig.~\ref{fig:app-controls}a), which drops the assistant prefill and decodes freely right after \texttt{Assistant:}, and 16-way scoring, which takes the softmax probability over summed log-likelihoods of the gold answer among 16 length-matched candidates of the same attribute (chance $6.25\%$).

We define the \emph{cue effect} for a given model, list length $N$, and cue condition as the recall accuracy with the cue minus the recall accuracy without it, in percentage points. The score is calculated using a paired evaluation: everything other than the cue (the document, ordering, target, question, answer prefill, and 16 candidates) is held identical, and we average the per-question differences. Since contexts within the same set of people share profiles, the experiments are not totally independent, and thus all 95\% confidence intervals are percentile bootstraps over the eight independent sets of biographies (10{,}000 resamples of per-set means).

\paragraph{Fine-tuning corpus and hyperparameters.}
We construct 4{,}096 profiles for fine-tuning using the same generation process (256 held out for validation). During fine-tuning, each model is exposed to 12{,}000 episodes of $N\in\{1,2,4,\dots,256\}$ biographies, while $N=512$ is held out as a length generalization evaluation. Every training cue is truthful and non-adversarial, and there are no uncued episodes. The next-token prediction cross-entropy loss is only computed on answer tokens (including end-of-text). For RWKV-7 World 2.9B, we also tried training with 30\% uncued episodes, which gave nearly identical uncued and cued recall and is omitted.

Table~\ref{tab:lm-ft-hparams} describes the hyperparameters used in fine-tuning. We found that using fp32 master weights was essential for performance: at learning rate $10^{-5}$, most bf16 updates round to zero (for instance, only 3.6\% of parameters in 1.3B Gated DeltaNet could move at all). G1i 13.3B is trained with FSDP so its sharded fp32 parameters serve as master weights, and G1i 7.2B is not fine-tuned. RWKV-7 World 2.9B, Gated DeltaNet, and DeltaNet are fine-tuned twice with independent seeds, and the other models once. All fine-tuned checkpoints are evaluated with the same protocol as above.

\begin{table}[ht]
\centering
\small
\caption{\textbf{Language-model fine-tuning hyperparameters} (Sec.~\ref{sec:real-models}).}
\label{tab:lm-ft-hparams}
\begin{tabular}{ll}
\toprule
Hyperparameter & Value \\
\midrule
Training episodes & 12{,}000 ($N\in\{1,2,4,\dots,256\}$; 512 held out) \\
Epochs / optimizer steps & 1 epoch / 187 steps (batch size 64: 8 GPUs $\times$ grad accum 8) \\
Optimizer & AdamW ($\beta=(0.9,0.95)$, $\epsilon=10^{-8}$, weight decay $0.01$) \\
Learning rate schedule & Peak $10^{-5}$, 3\% linear warmup, cosine decay to 0 \\
Gradient clipping & 1.0 \\
Precision / master weights & bfloat16 forward/backward, fp32 master weights (FSDP for 13.3B) \\
Max sequence length & 22{,}528 tokens (RWKV-7) / 28{,}672 tokens (1.3B models, Pythia) \\
\bottomrule
\end{tabular}
\end{table}

\subsection{Additional results}
\label{app:roster}

\begin{figure}[!htbp]
    \centering
    \includegraphics[width=0.85\textwidth]{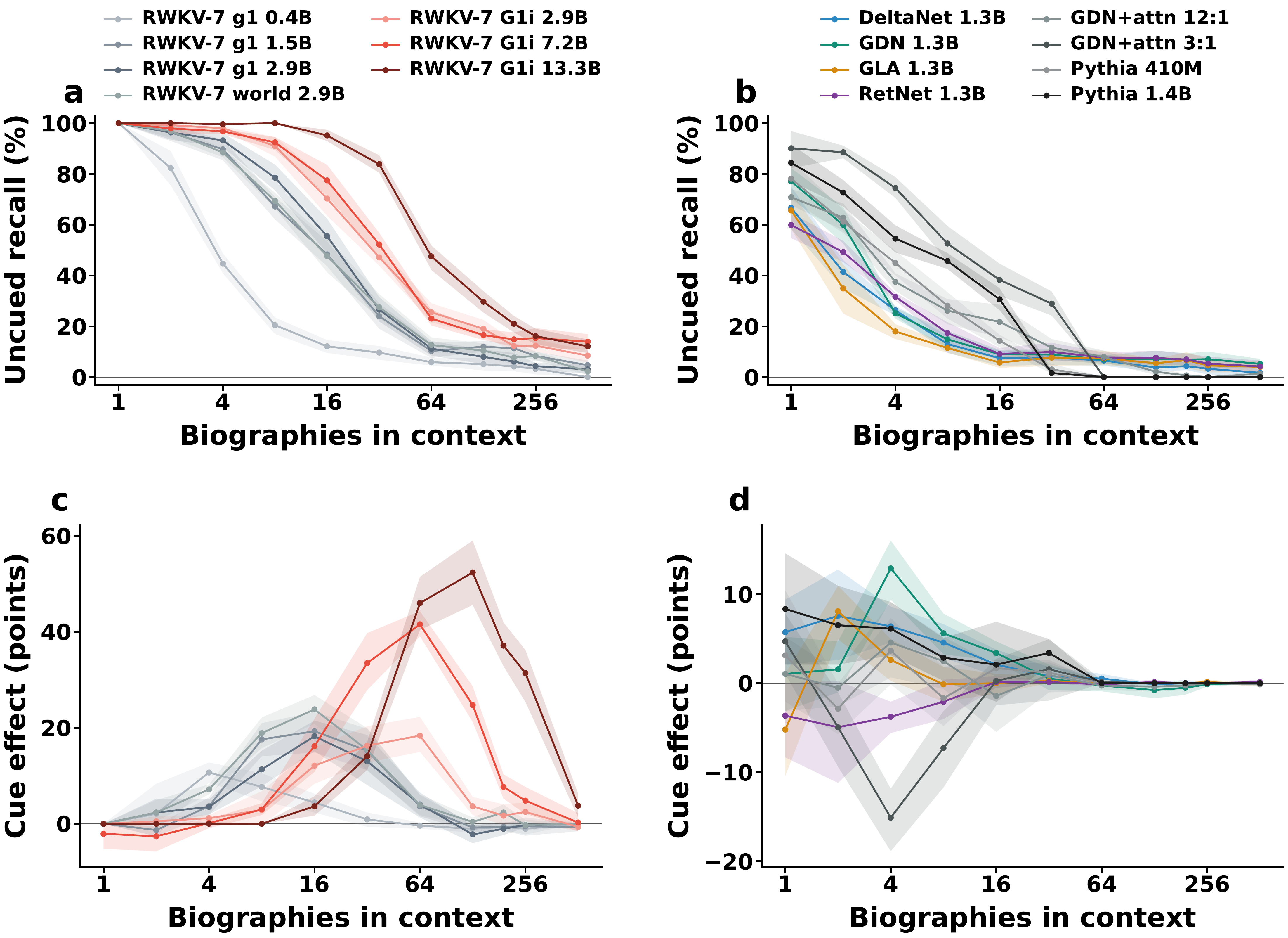}
    \caption{\textbf{Every pretrained model.}
    \textbf{a, b.} Uncued recall for the RWKV-7 checkpoints (a) and the other architectures (b).
    \textbf{c, d.} Exact-name cue effect (remember A minus no cue) for the same models.
    Shading: 95\% bootstrap intervals over eight sets of people.}
    \label{fig:app-roster}
\end{figure}

\paragraph{Every pretrained model.}
In Fig.~\ref{fig:app-roster}, we show all pretrained models across all lengths, extending Fig.~\ref{fig:real-pretrained}. Note that for RWKV-7, the cue effect grows with scale \emph{only} in the G1i line; the g1 and World models peak at 11--24 points between 4 and 16 biographies. The other architectures show effects of either sign up to about 15 points at short lengths, and none shows an effect beyond 64 biographies.

\begin{figure}[!htbp]
    \centering
    \includegraphics[width=0.85\textwidth]{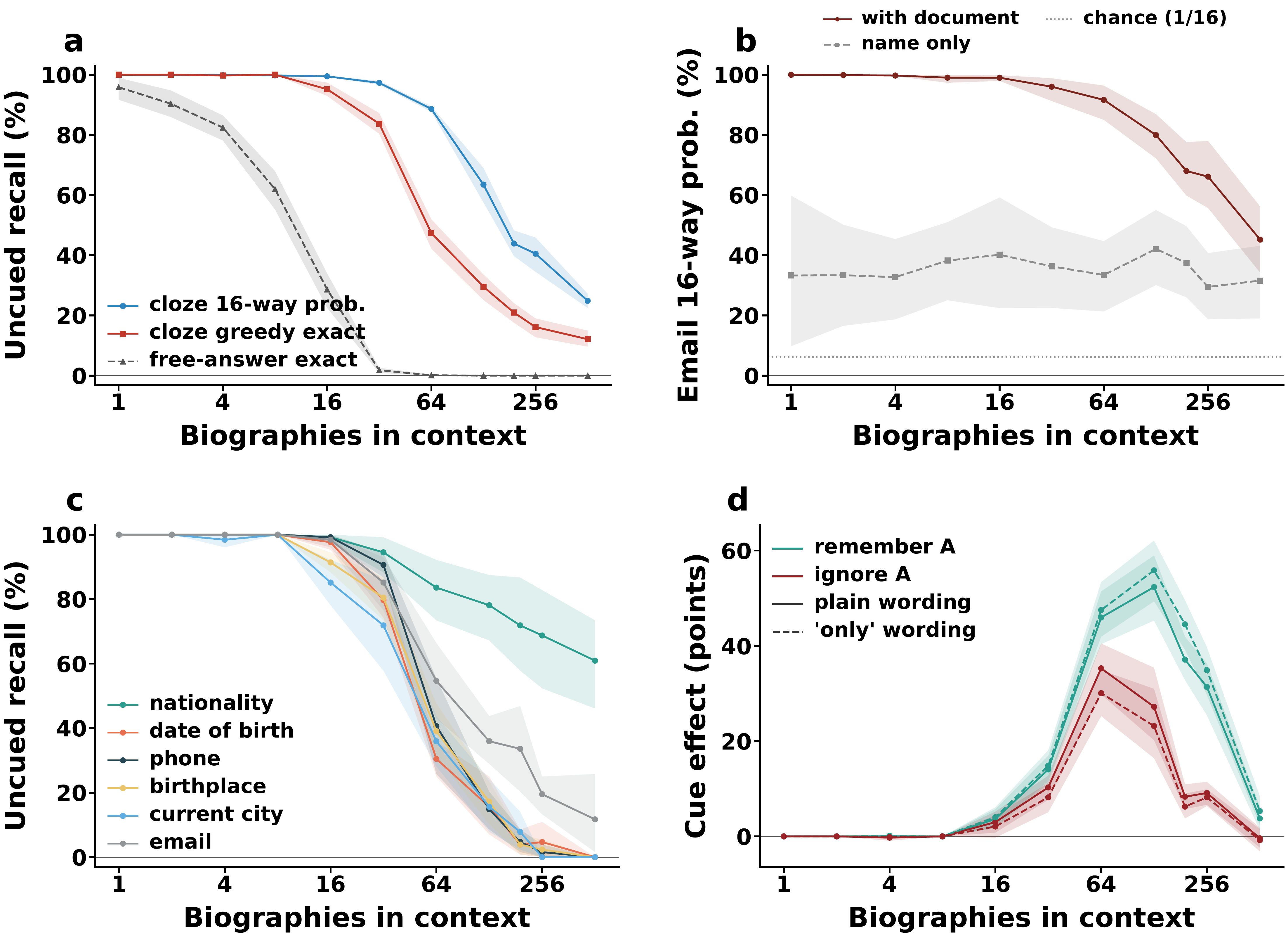}
    \caption{\textbf{Scoring and measurement controls (RWKV-7 G1i 13.3B, uncued unless noted).}
\textbf{a.} Answer format: probability of the gold answer among 16 candidates, greedy exact match of the cloze answer (used in the main text), and exact match of a free answer. Free answers fail first, and the main-text cloze scoring lies between the two.
\textbf{b.} Hard email control: 16-way probability of the gold email when all 15 alternatives are addresses built from the same person's name, with the biographies (solid) or with the name alone (dashed). Without the document the model still reaches about 35\%, far above chance (dotted, 1/16), so part of email recall is guessing from the name.
\textbf{c.} Recall by queried attribute. Nationality stays easiest at long contexts.
\textbf{d.} Cue wording: ``only remember A'' (dashed) behaves like the plain ``remember A'' wording (solid), for both the remember and the ignore cue.
Shading: 95\% bootstrap intervals over eight sets of people.}
    \label{fig:app-controls}
\end{figure}

\begin{figure}[!htbp]
    \centering
    \includegraphics[width=0.85\textwidth]{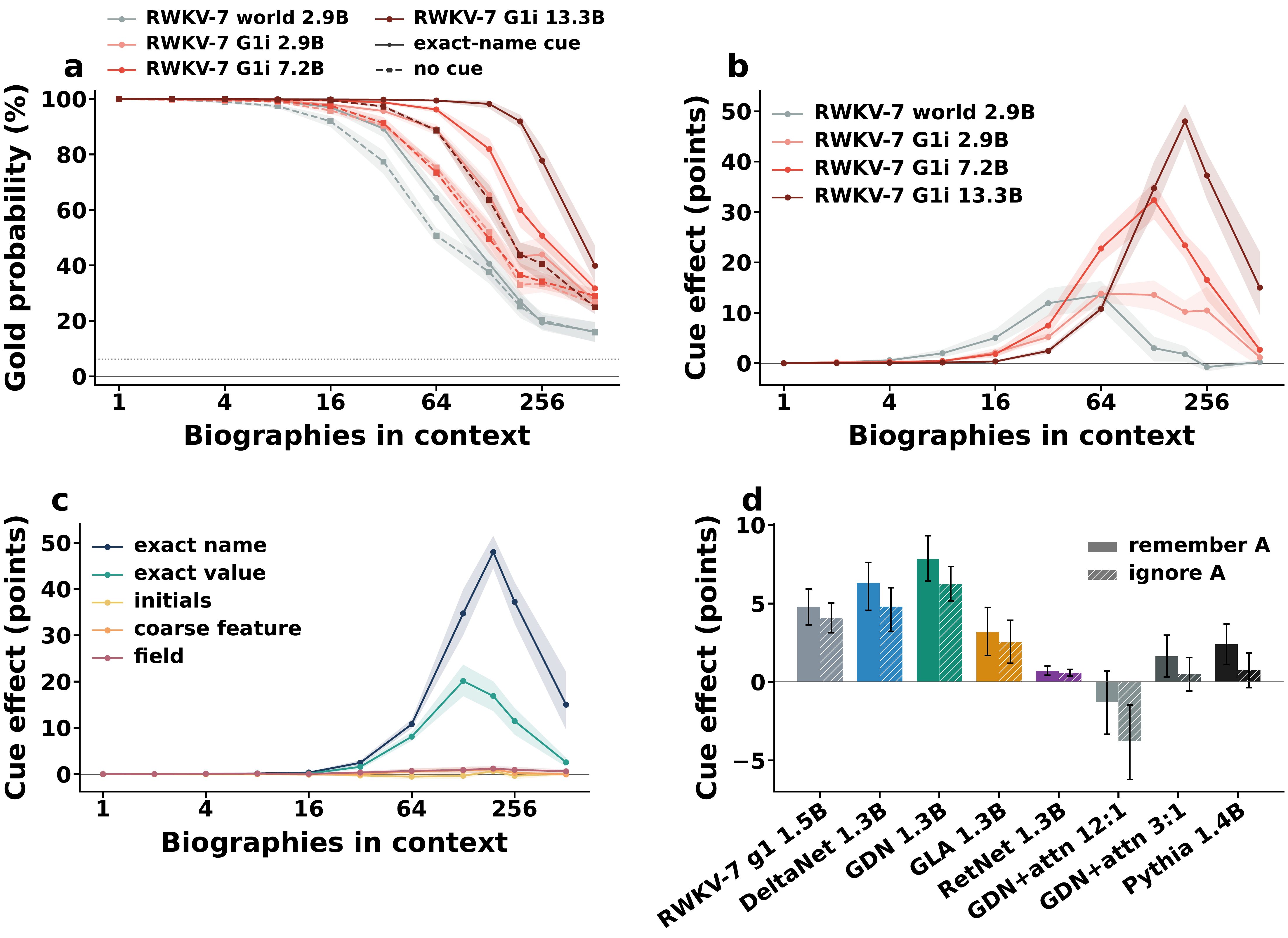}
    \caption{\textbf{Fig.~\ref{fig:real-pretrained} with 16-way scoring.}
Recall is the probability of the gold answer among 16 candidates (chance: 6.25\%, dotted in a).
\textbf{a.} Recall with (solid) and without (dashed) an exact-name cue for four RWKV-7 checkpoints (Fig.~\ref{fig:real-pretrained}b).
\textbf{b.} Exact-name cue effect across scale (Fig.~\ref{fig:real-pretrained}c).
\textbf{c.} Cue categories for G1i 13.3B (Fig.~\ref{fig:real-pretrained}d).
\textbf{d.} Remember-A (solid) and ignore-A (hatched) cue effects across architectures at 16 biographies (Fig.~\ref{fig:real-pretrained}f).
Shading and error bars: 95\% bootstrap intervals over eight sets of people.}
    \label{fig:app-16way}
\end{figure}

\paragraph{Scoring and measurement controls.}
\label{app:bio-controls}
In Sec.~\ref{sec:real-models}, we reported recall using greedy exact match. Figs.~\ref{fig:app-controls} and~\ref{fig:app-16way} check this choice on RWKV-7 G1i 13.3B and repeat Fig.~\ref{fig:real-pretrained}b--d and~f with 16-way scoring: the scale trend and cue categories are reproduced, with the exact-name cue effect reaching 48 points for G1i 13.3B and only exact text helping. The main difference under 16-way scoring is that at 16 biographies every architecture's effect is smaller (below 10 points), with DeltaNet (6 points) and Gated DeltaNet (8 points) slightly above RWKV-7 g1 1.5B (5 points; Fig.~\ref{fig:app-16way}d).

\begin{figure}[!htbp]
    \centering
    \includegraphics[width=\textwidth]{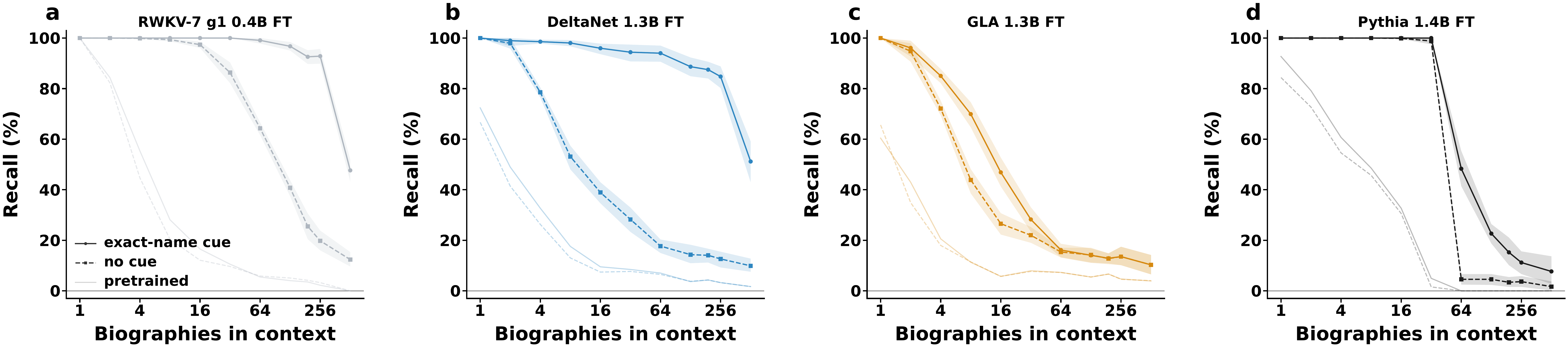}
    \caption{\textbf{Fine-tuned models, one per outcome.}
Recall with (solid) and without (dashed) an exact-name cue after fine-tuning; faint lines: the same model before fine-tuning.
\textbf{a.} The smallest RWKV-7 (g1 0.4B) learns strong uncued and cued recall and keeps cued recall above 90\% up to 256 biographies.
\textbf{b.} DeltaNet recalls well only what it was cued to keep.
\textbf{c.} GLA gains little from the cue.
\textbf{d.} The Transformer Pythia 1.4B recalls nearly everything up to 32 biographies (about 2{,}500 tokens) and fails from 64 on, although it was fine-tuned on up to 256 biographies; its pretraining context is 2{,}048 tokens.
One training seed per model; shading: 95\% bootstrap intervals over eight sets of people.}
    \label{fig:app-ft-models}
\end{figure}

\begin{figure}[!htbp]
    \centering
    \includegraphics[width=\textwidth]{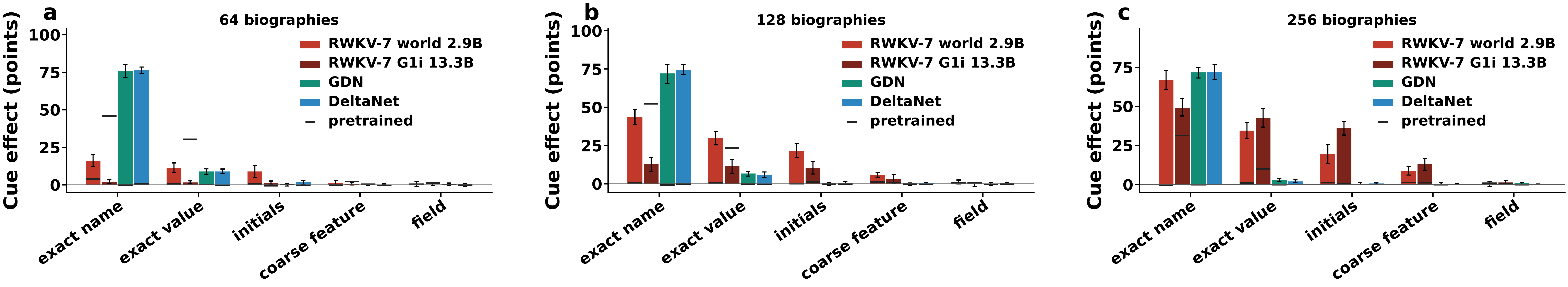}
    \caption{\textbf{Cue categories after fine-tuning at 64 (a), 128 (b) and 256 (c) biographies}
(Fig.~\ref{fig:real-finetuned}b is 256). Black ticks: the same model before fine-tuning.
At shorter lengths the fine-tuned RWKV-7 models still recall most records without a cue, which caps their cue effects, while Gated DeltaNet and DeltaNet already gain 70--75 points from the exact name at 64. The ordering of the categories is the same at every length.
Error bars: 95\% bootstrap intervals over eight sets of people.}
    \label{fig:app-ft-lengths}
\end{figure}

\begin{figure}[!htbp]
    \centering
    \includegraphics[width=\textwidth]{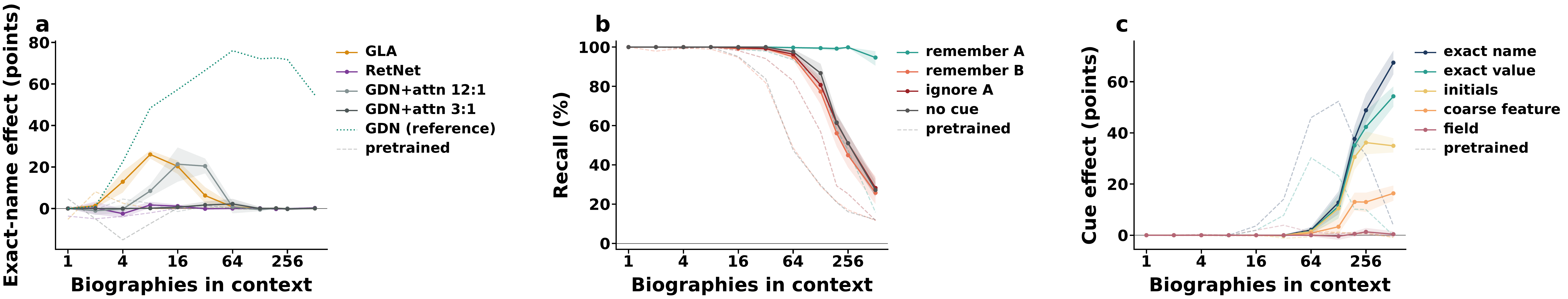}
    \caption{\textbf{Short-range cue use, and fine-tuning at 13.3B.}
\textbf{a.} After fine-tuning, GLA and the 12:1 Gated DeltaNet--attention hybrid gain from the exact-name cue only at short lengths (at most 26 and 21 points, between 8 and 32 biographies), and RetNet and the 3:1 hybrid hardly at all. Dotted: fine-tuned Gated DeltaNet for reference; faint dashed lines: before fine-tuning.
\textbf{b.} RWKV-7 G1i 13.3B after fine-tuning: recall under remember A, remember B, ignore A and no cue; faint dashed lines: before fine-tuning. ``Ignore A'' no longer helps.
\textbf{c.} Cue categories for the same model over every length.
One training seed; shading: 95\% bootstrap intervals over eight sets of people.}
    \label{fig:app-ft-short}
\end{figure}

\paragraph{Fine-tuned models.}
\label{app:bio-ft}
While Fig.~\ref{fig:real-finetuned} shows World 2.9B, Gated DeltaNet, and DeltaNet, Fig.~\ref{fig:app-ft-models} shows fine-tuning results for the other models (RWKV-7 g1 0.4B, DeltaNet, GLA, and Pythia 1.4B). Fig.~\ref{fig:app-ft-lengths} repeats the cue-category comparison across 64, 128, and 256 biographies. Finally, Fig.~\ref{fig:app-ft-short} shows the architectures that only use the cue at short lengths alongside the largest fine-tuned model, G1i 13.3B, across all lengths.

\clearpage
\section{Synthetic key--value recall task}
\label{app:exp-synthetic}

This section provides experimental details and additional results for the from-scratch synthetic key--value recall experiments in Sec.~\ref{sec:synthetic} (Fig.~\ref{fig:synthetic}).

\subsection{Experimental details}
\label{app:synthetic-setup}

\paragraph{Episodes.}
The vocabulary has 4{,}096 key tokens, 256 value tokens and a few control tokens, all of which are separate single tokens. Each episode has three parts:
\begin{enumerate}
    \item \texttt{BOS} and ten cue slots, each \texttt{PRIME} followed by a key or \texttt{NULL}.
    \item \texttt{DOC} and $N$ key--value records.\footnote{During training, each record also carries a \texttt{MARK} or \texttt{NOMARK} tag (with $M\sim\mathcal{U}\{10,\dots,N\}$ marked records from which the ten questions are drawn). All reported evaluations mark every record ($M=N$), so the tag carries no information.}
    \item Ten questions, each written as \texttt{ASK} key \texttt{ANS} value.
\end{enumerate}
The keys in the context are sampled without replacement and the values are sampled uniformly at random. Ten records are selected as questions, and $K$ of the ten question keys are placed in the cue slots in random order. The remaining cue slots are simply filled with a \texttt{NULL} token so the sequence length does not depend on $K$. A question is \emph{cued} if its key appeared in one of the cue slots and \emph{uncued} otherwise.

\paragraph{Models and training.}
Unless otherwise mentioned, models have 4 layers, hidden size $d=256$ and MLP ratio 4. Table~\ref{tab:synth-hparams} lists the detailed architectural and training hyperparameters. Our Transformers include Canon layers \citep{allenzhu2026physicslanguagemodels41} (short local convolutions added to the block; Canon-ACD with kernel size 4 for the main Transformers, and Canon-B with kernel size 2 for the KV-budget Transformer below). Note that the Gated DeltaNet in Figs.~\ref{fig:synthetic} and~\ref{fig:app-architectures} uses 12 heads of dimension 16 with value expansion 2 and a width-4 convolution (24{,}576 matrix floats, 6.20M parameters). Thus its state is larger than that of RWKV-7, which has 16{,}384 matrix floats. For the hybrid models in Fig.~\ref{fig:app-mechanism}c, every layer uses the same feed-forward block (RMSNorm, residual and SwiGLU) and only swaps the sequence mixer between RWKV-7 (R) and Gated DeltaNet (G).

\begin{table}[ht]
\centering
\footnotesize
\caption{\textbf{Synthetic key--value recall architectures and training hyperparameters} (Sec.~\ref{sec:synthetic}).}
\label{tab:synth-hparams}
\begin{tabular}{lp{0.66\linewidth}}
\toprule
Setting & Value \\
\midrule
RWKV-7 (main) & 4 layers, $d=256$, 16 heads ($d_h=16$), 16{,}384 matrix floats, 5.95M params \\
RWKV-7 width sweep & $d=128$ (8 heads, 8{,}192 floats, 2.19M) and $d=512$ (16 heads, 65{,}536 floats, 18.19M) \\
Gated DeltaNet (Fig.~\ref{fig:synthetic}) & 4 layers, $d=256$, 12 heads ($d_h=16$), val exp 2, conv 4, 24{,}576 floats, 6.20M params \\
Transformer + Canon & 4 layers, $d=256$, 4 heads, RoPE ($\theta=10^4$) or NoPE, Canon ACD ($k=4$), 6.47M params \\
GRU / simple RNN & $d=352$ (6.06M) / $d=480$ (6.04M), Canon layer A \\
Sampling per batch & $N\sim\mathcal{U}\{32,64,\dots,N_{\max}\}$ ($N_{\max}\in\{128,256,512,1024\}$), $K\sim\mathcal{U}\{0,\dots,10\}$ \\
Optimization & 100k steps, batch 32, AdamW ($\beta=(0.9,0.95)$, wd $0.01$), peak lr $3\times10^{-4}$ \\
Schedule & 2.5\% linear warmup, linear decay to 0, grad clip 1.0, loss on answer tokens only \\
\bottomrule
\end{tabular}
\end{table}

\paragraph{Evaluation and statistical notes.}
The model reads each document once and answers every question greedily from its own copy of the post-document state; as in App.~\ref{app:lm-eval}, cued and uncued questions are evaluated on the same paired document context. We evaluate $N\in\{32,48,\dots,4096\}$ (15 lengths) and $K\in\{0,2,5,8,10\}$, with 16 documents per setting. Reported figures pool results over $K$ and report exact-match accuracy on the value token (chance $1/256$). Error bars in Fig.~\ref{fig:synthetic} are standard errors of the mean over 128 evaluation documents for one trained model, while Fig.~\ref{fig:app-acquisition} runs three seeds and plots each seed separately.

\paragraph{Multi-seed sweep with equal state size (Fig.~\ref{fig:app-acquisition}).}
For this experiment, every recurrent model has 16{,}384 matrix floats. Gated DeltaNet uses 16 heads of dimension 16, value expansion 1 and a width-2 convolution, and DeltaNet uses 16 heads and a width-2 convolution.

The warm-started Gated DeltaNet is trained in two stages:
\begin{enumerate}
    \item Training on shorter lists ($N\le256$) for 100k steps at lr $10^{-4}$.
    \item Continued training from that checkpoint (same seed) on $N\le1024$ for 100k steps at lr $10^{-4}$.
\end{enumerate}

Each of the other variants changes one setting:
\begin{itemize}
    \item \texttt{long300k} trains for 300k steps.
    \item \texttt{lr1e4} and \texttt{lr2e4} change the learning rate.
    \item \texttt{boost} oversamples $N\in\{32,64\}$.
    \item \texttt{kall} reveals every cued key.
    \item \texttt{dtinit} re-initializes the range of the decay gate.
    \item \texttt{warmbeta}, \texttt{warmgate} and \texttt{warmgates} copy over only part of the first-stage checkpoint.
    \item \texttt{noconv} removes the short convolution.
    \item \texttt{rwshape} lays out the decay the way RWKV-7 does.
\end{itemize}

We also include a Transformer with a limited key--value budget (4 heads, Canon-B with kernel size 2), trained with $N\le256$. The budget works as follows: for each document token, the Transformer learns an internal score that decides whether the token is kept or dropped, and attention is restricted to the top $B\in\{192,384\}$ document tokens seen so far. The start, cue and query tokens are always attended to.

Sweep runs are evaluated on 3{,}600 documents (36{,}000 questions) for $N_{\max}=1024$ and on 3{,}120 documents for $N_{\max}=256$. To measure how quickly each run learns to use the cue (Fig.~\ref{fig:app-acquisition}c), we run a quick evaluation (2 documents per length) every 2{,}500 training steps and record the first step where cued minus uncued accuracy stays $\ge0.1$ for three consecutive checks. Warm-started runs are excluded since they already use the cue at step 0.

\subsection{Additional results}
\label{app:formula-sweep}

\begin{figure}[!htbp]
    \centering
    \includegraphics[width=0.85\textwidth]{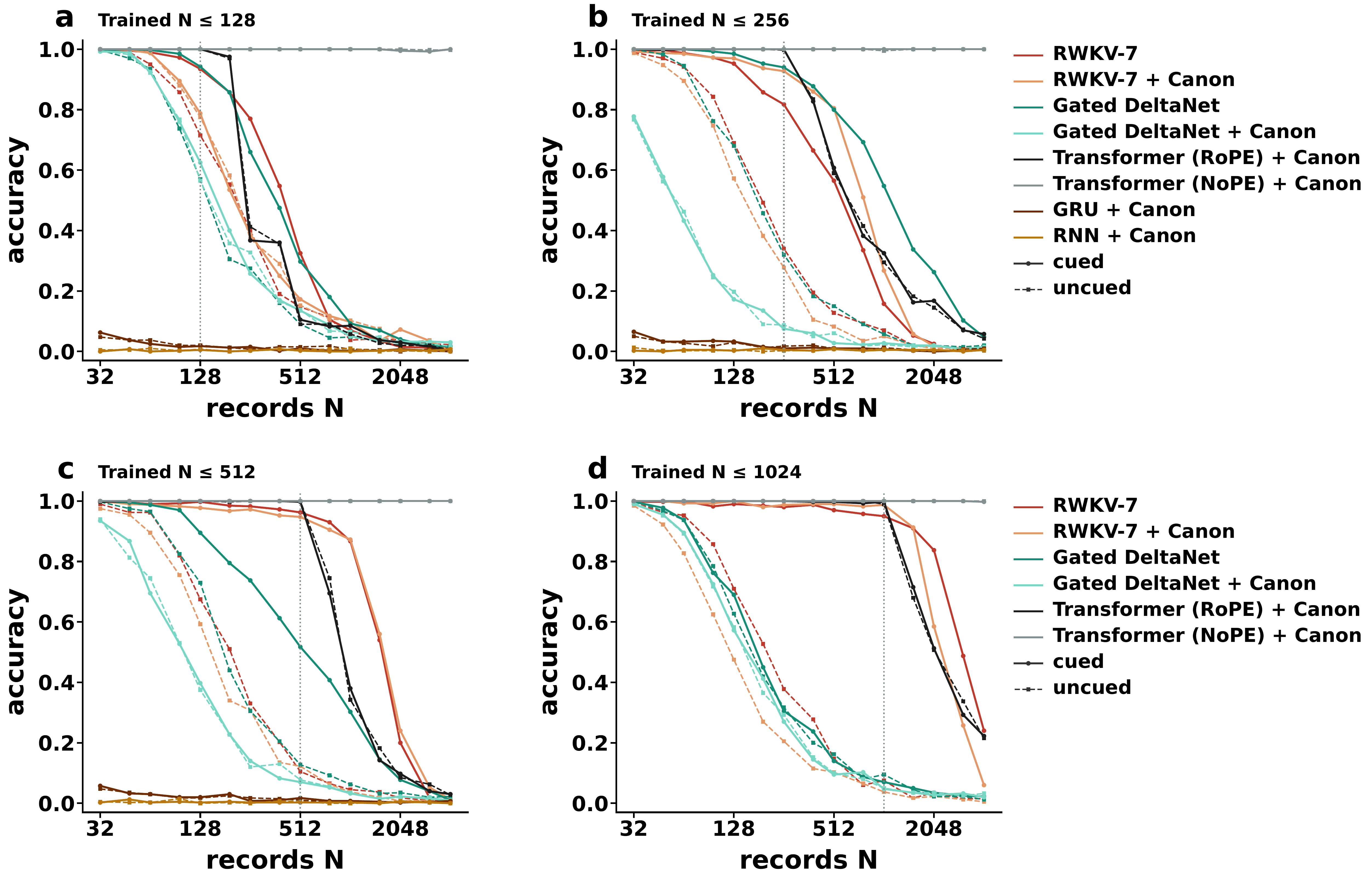}
    \caption{\textbf{Every architecture, by training length.}
Cued (solid) and uncued (dashed) accuracy against the number of records $N$ (pooled over $K$), for models trained
with $N\le128$ (a), $256$ (b), $512$ (c) and $1024$ (d); dotted vertical line: the training limit.
``+ Canon'': with Canon layers. GRU and RNN have no $N\le1024$ run. One training seed per model.}
    \label{fig:app-architectures}
\end{figure}

\paragraph{Architectures and training length.}
Fig.~\ref{fig:app-architectures} shows cued and uncued accuracy for all architectures trained up to four maximum lengths (one seed each). The NoPE Transformer \citep{kazemnejad2023impact} solves both cued and uncued questions at every length and thus has no need for the cue. The RoPE Transformer \citep{su2023roformerenhancedtransformerrotary} fails to length-generalize beyond its training length whether or not it is cued. The GRU and simple RNN baselines fail to learn the base task within a reasonable gradient budget.

\paragraph{How often and how quickly models learn to use the cue.}
\label{app:arch-budget}

\begin{figure}[!htbp]
    \centering
    \includegraphics[width=\textwidth]{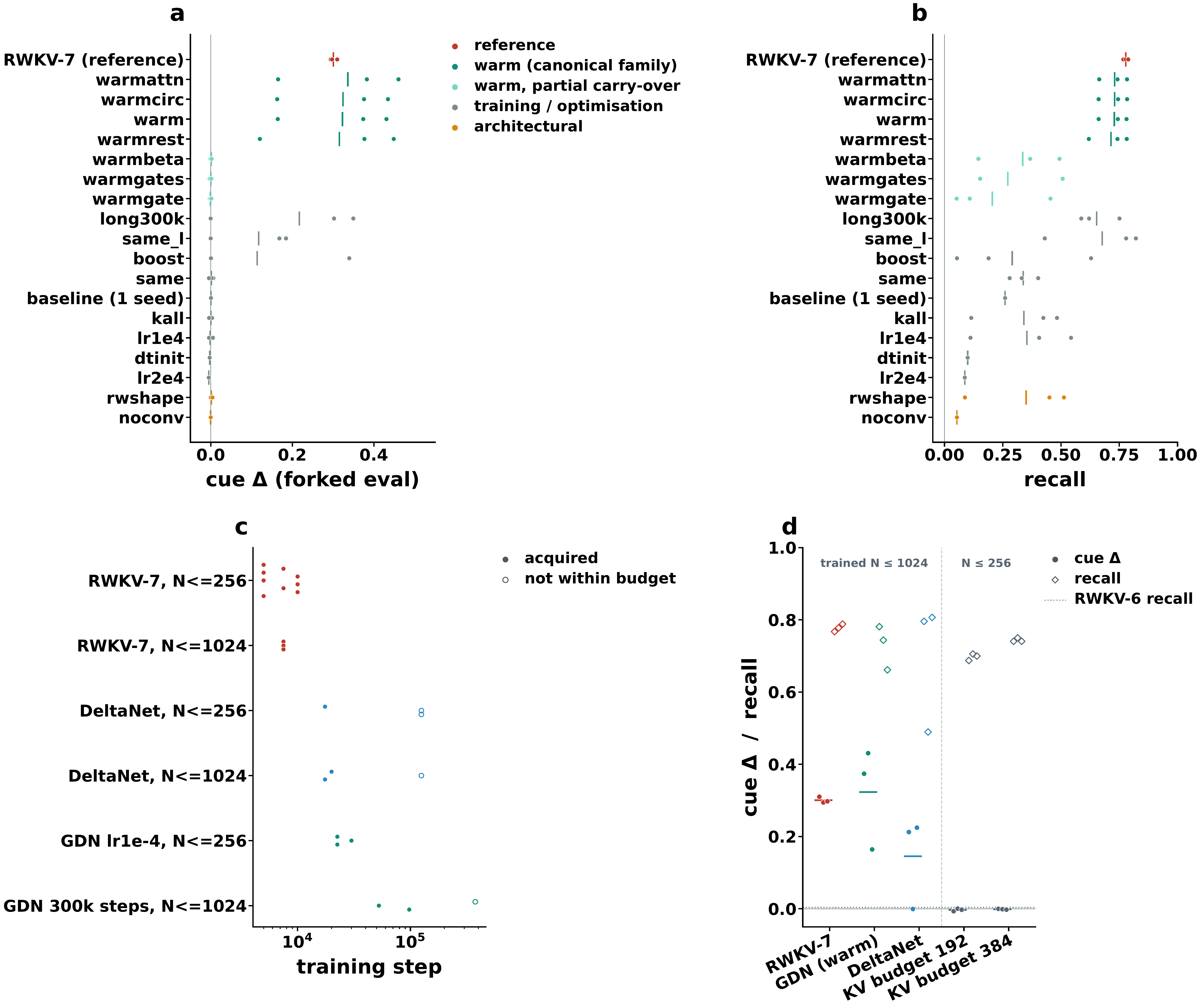}
    \caption{\textbf{Acquisition rate and training budget (state-matched: 16{,}384 recurrent-matrix floats).}
\textbf{a, b.} Gated DeltaNet variants trained through $N=1024$ (App.~\ref{app:synthetic-setup}): cue advantage (a) and recall (b)
on the 36k-question evaluation; dots: seeds (three, except one for the baseline, \texttt{dtinit} and \texttt{lr2e4}); bars: means; RWKV-7 as reference; colours group the variants.
\textbf{c.} Training step at which each run first meets the acquisition criterion (App.~\ref{app:synthetic-setup}); open
circles: not met within the budget. RWKV-7 with $N\le256$ has nine seeds, every other arm three.
\textbf{d.} Cue advantage (dots, bar: mean) and recall (diamonds) by architecture; KV-budget Transformers are trained with $N\le256$, the others with $N\le1024$;
dotted: RWKV-6 recall, which never learns the base task.}
    \label{fig:app-acquisition}
\end{figure}

In Fig.~\ref{fig:app-acquisition}, every recurrent model has the same state size of 16{,}384 floats and each model is trained with three seeds. RWKV-7 learns to use the cue in every seed. Gated DeltaNet learns it in all seeds \emph{only} when warm-started from the full first-stage checkpoint described in App.~\ref{app:synthetic-setup}; more broadly, Gated DeltaNet shows much higher seed sensitivity when training for cued recall. Across architectures, the step at which a run learns to use the cue varies by more than an order of magnitude. Transformers with a limited key--value budget ($B\in\{192,384\}$, $N\le256$) reach 70--75\% recall but show no cue effect.

\paragraph{Width tradeoff.}
\label{app:width}

\begin{figure}[!htbp]
    \centering
    \includegraphics[width=0.85\textwidth]{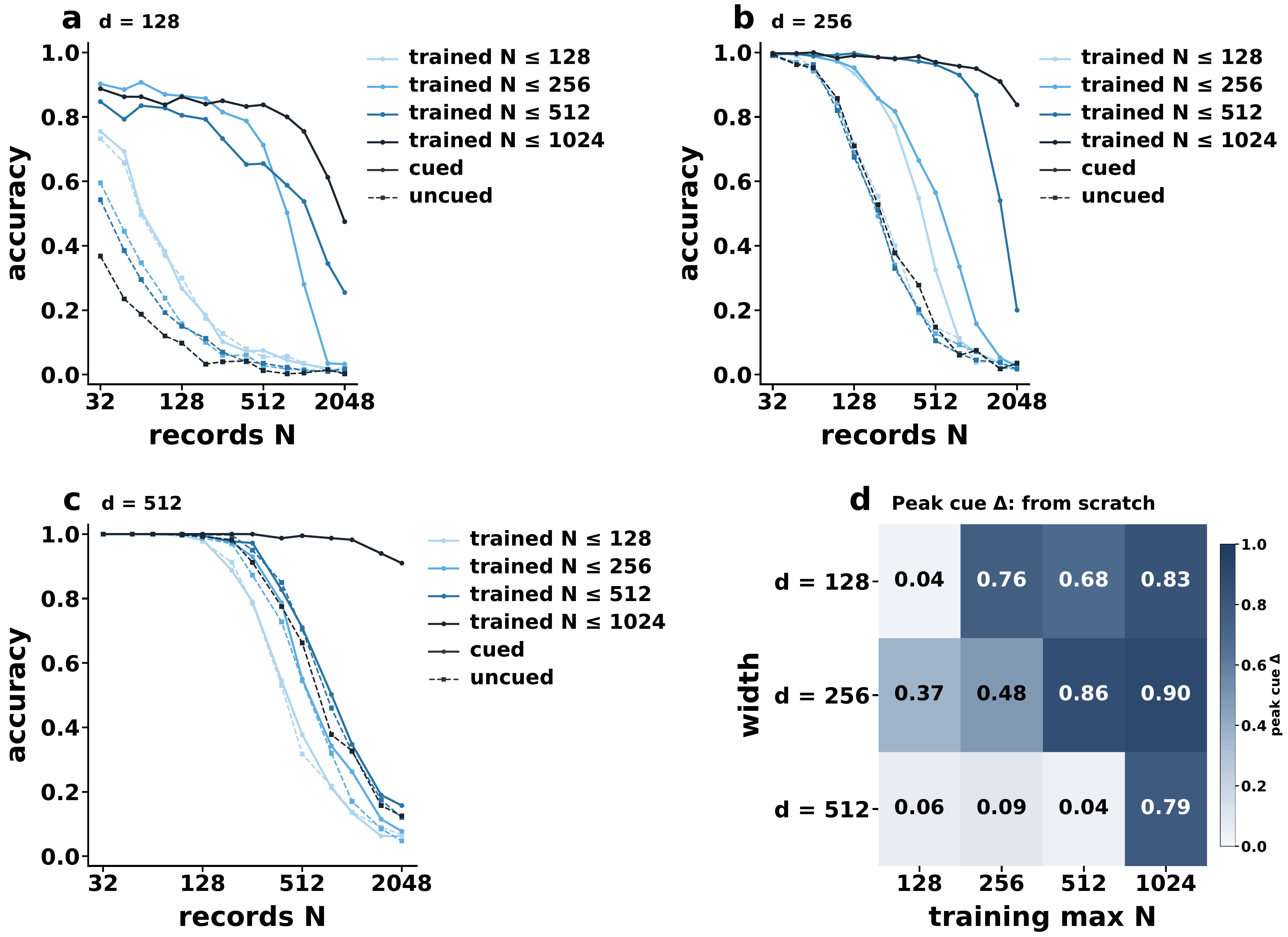}
    \caption{\textbf{Width tradeoff.}
\textbf{a--c.} Cued (solid) and uncued (dashed) accuracy for RWKV-7 with hidden size $d=128$, $256$ and
$512$ (8{,}192, 16{,}384 and 65{,}536 matrix floats), one line per training length.
\textbf{d.} Peak cue effect, the maximum over evaluated $N$ of cued minus uncued accuracy, by width and training length (trained from scratch).
Width also changes the parameter count (2.2M, 5.9M, 18.2M). One training seed per cell.}
    \label{fig:app-width}
\end{figure}

Fig.~\ref{fig:app-width} shows the full width sweep behind Fig.~\ref{fig:synthetic}d across training lengths. We find that, as expected, wider states keep uncued recall high up to longer lists. Learning of the cue depends heavily on the training range: while $d=512$ trained up to $N=512$ peaks at a cue effect of only 0.04, the same width trained up to $N=1024$ reaches 0.79.

\paragraph{Where the cue is stored, how it changes what is written and the effect of list position.}
\label{app:causal}

\begin{figure}[!htbp]
    \centering
    \includegraphics[width=0.85\textwidth]{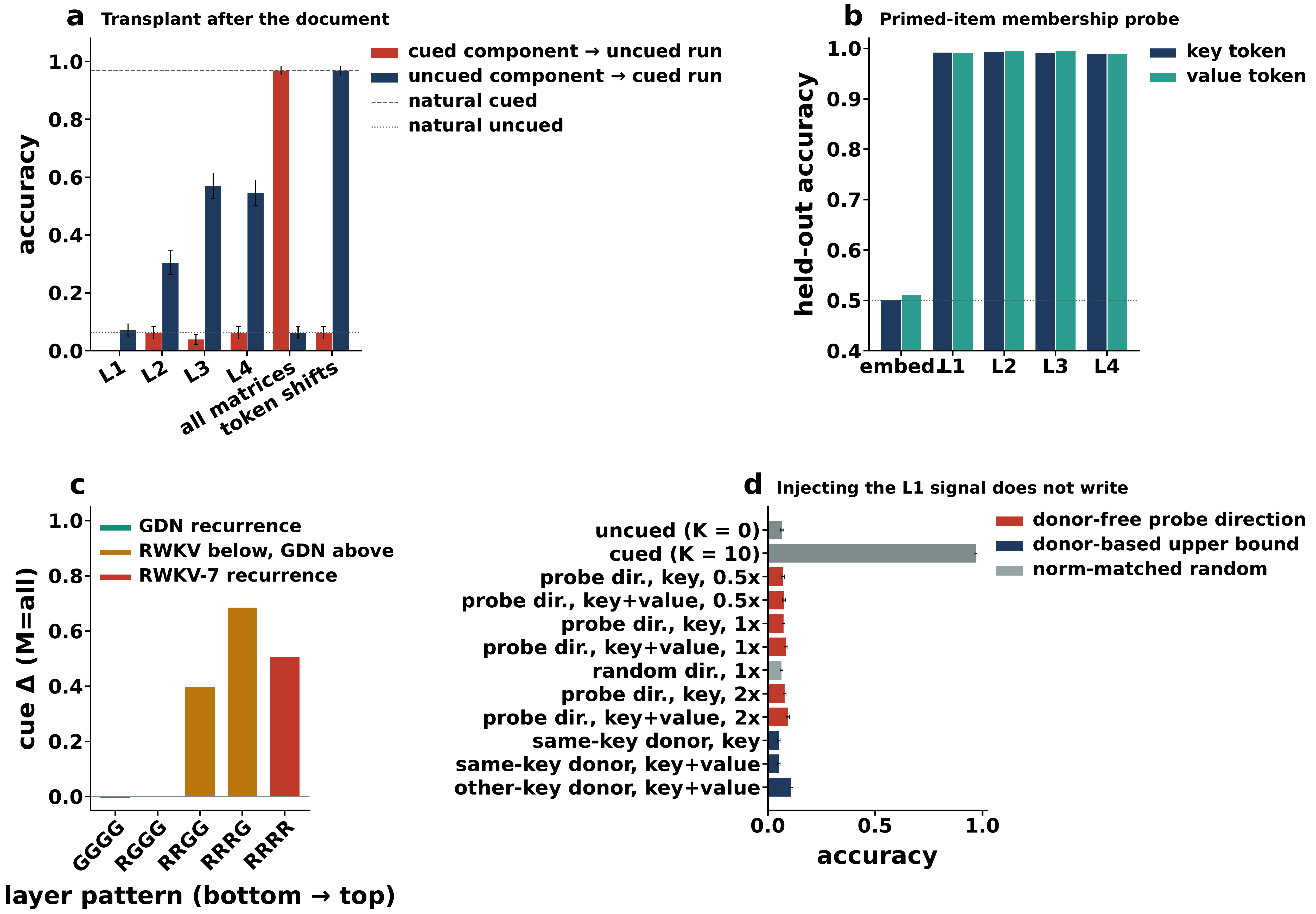}
    \caption{\textbf{Where the cue is held} (RWKV-7 trained with $N\le1024$, evaluated at $N=1024$).
\textbf{a.} Transplants after the document (128 documents, one cued question each): copying a component of the cued state into the
uncued run (red), or of the uncued state into the cued run (navy). Moving all matrix states
transfers cued recall and no single layer does (the cued layer 1 alone gives 0 of 128); replacing layer 1 with its uncued state removes cued recall (0.07), and replacing any of layers 2--4 lowers it (0.30--0.57).
\textbf{b.} A linear probe reads from each layer's state whether a key or value was primed ($K=10$);
embeddings are at chance.
\textbf{c.} Cue effect for four-layer hybrids, listed bottom to top (R: RWKV-7, G: Gated DeltaNet).
The policy appears once the two lowest layers are RWKV-7.
\textbf{d.} With no cue, adding the probe direction for each queried key (and value) to the layer-1 output at three
scales leaves accuracy at the uncued level. Adding donor states at the same site, and a norm-matched random direction, does no better.
One training seed; error bars: standard error over documents (a) or questions (d, 1{,}280).}
    \label{fig:app-mechanism}
\end{figure}

In order to understand where the cue is stored and how it acts, we transplant recurrent states from a cued run into an uncued run (and vice versa). Fig.~\ref{fig:synthetic}e transplants states right after the cue (before the document). Fig.~\ref{fig:app-mechanism} adds four further tests:
\begin{itemize}
    \item Transplanting states from after the document;
    \item Training linear probes for the cued keys;
    \item Mixing RWKV-7 and Gated DeltaNet layers in hybrids;
    \item Trying to reproduce cued recall without copying in a state from a cued run.
\end{itemize}

We find that transplanting a single layer after the document is not enough. Layer 1 is necessary and holds the cue itself (removing it drops cued recall to 0.07), while layers 2--4 hold the value content. We measure this by editing one value and splitting the resulting change in the answer logits across layers with Shapley values, giving 0.0, 9.2, 4.9 and 4.7 for layers 1--4. Only hybrids whose two lowest layers are RWKV-7 learn to use the cue, and adding the probe direction or a cued state at the output of layer 1 does not reproduce cued recall.

\begin{figure}[!htbp]
    \centering
    \includegraphics[width=\textwidth]{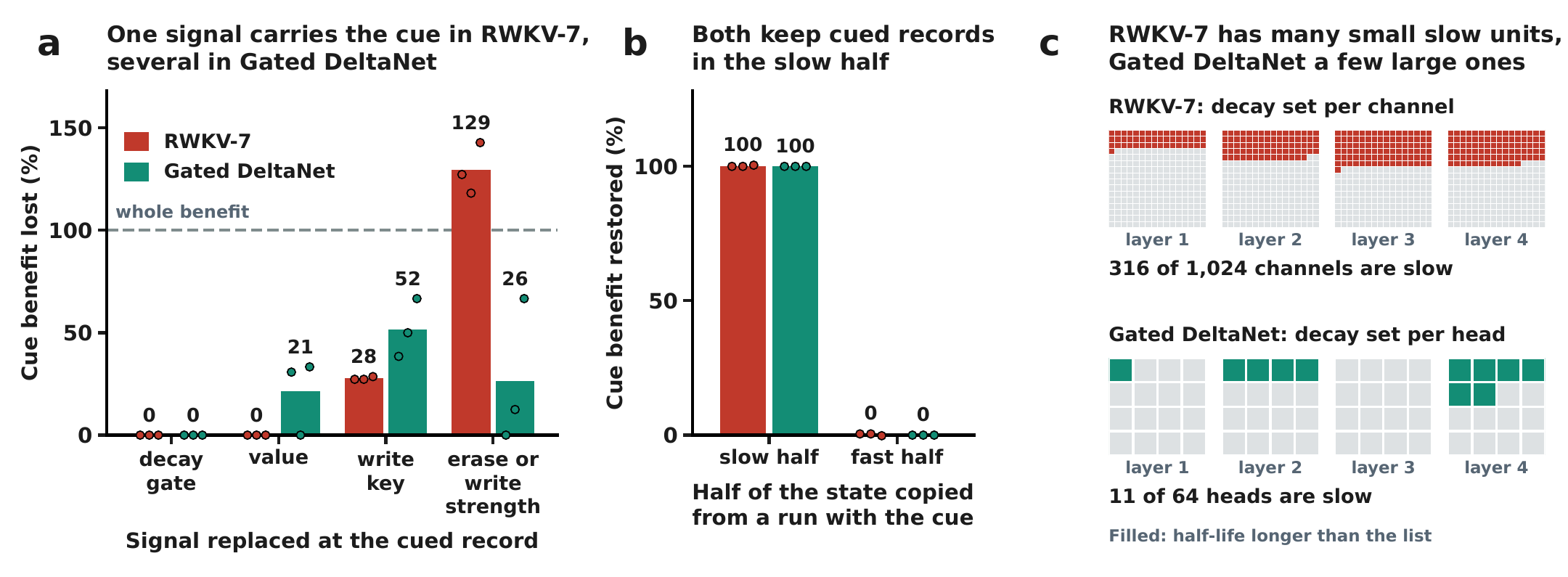}
    \caption{\textbf{What the cue does in RWKV-7 (red) and Gated DeltaNet (green).} \textbf{a.} One write signal at the cued record is replaced by its value from a run without the cue. Above 100\%, recall falls below the level without the cue. \textbf{b.} Half of the state is copied from a run with the cue into a run without it. \textbf{c.} The units that set decay in each layer, one cell per unit, for one run of each architecture. Cue benefit: recall with the cue minus recall without it. Bars: means over three seeds. Dots: seeds.}
    \label{fig:app-cue-write}
\end{figure}

\begin{figure}[!htbp]
    \centering
    \includegraphics[width=\textwidth]{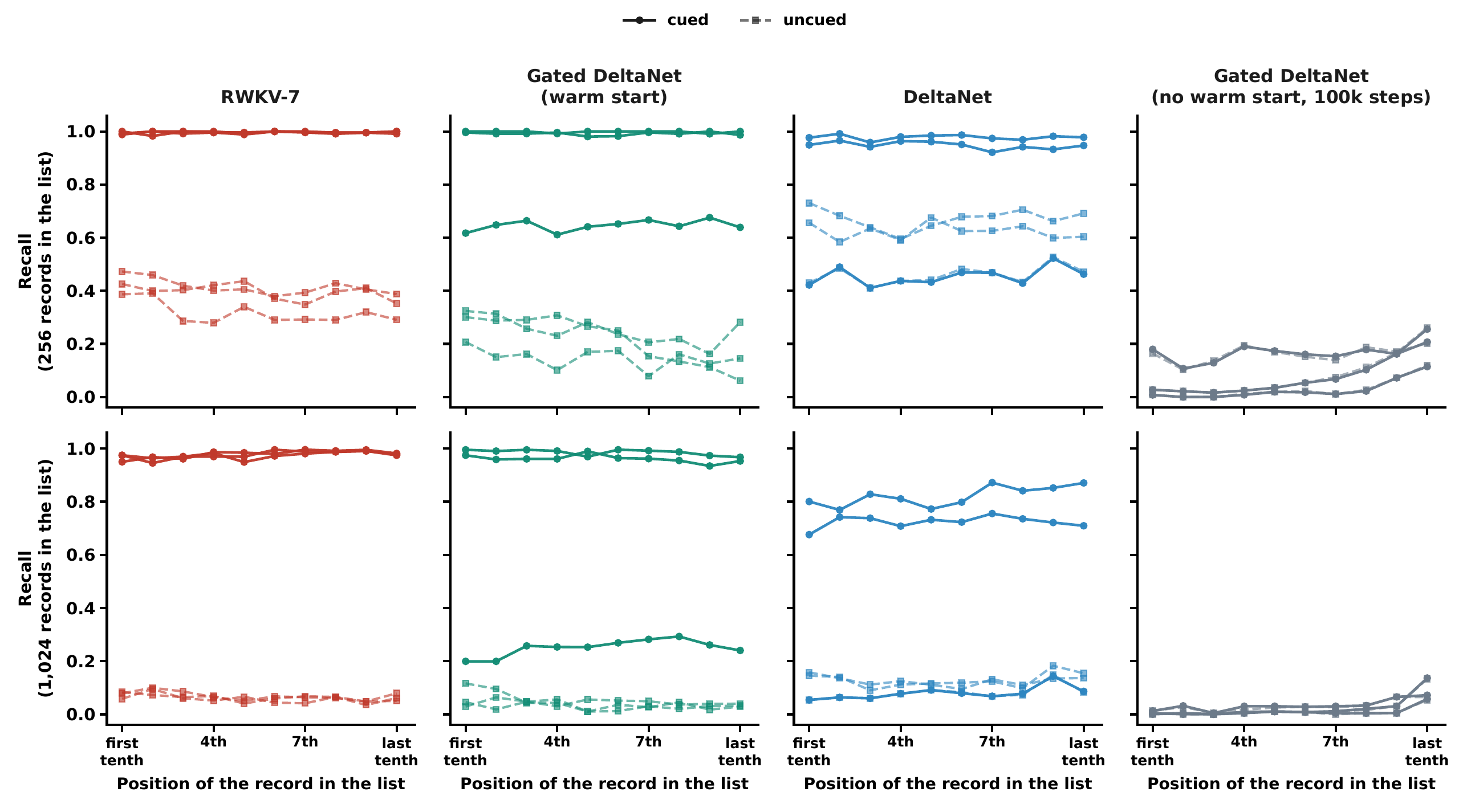}
    \caption{\textbf{Recall by position of the record in the list}, with the cue (solid) and without it (dashed), for lists of 256 records (top) and 1{,}024 records (bottom). One line per seed.}
    \label{fig:app-position}
\end{figure}

Figs.~\ref{fig:app-cue-write} and~\ref{fig:app-position} show results on three additional tests across three seeds:
\begin{enumerate}
    \item \textbf{Write-signal replacement (Fig.~\ref{fig:app-cue-write}A):} replacing a single write signal at the cued record with its value from an uncued run drops cued recall sharply.
    \item \textbf{Slow vs.\ fast state copying (Fig.~\ref{fig:app-cue-write}B,c):} copying the slowly decaying half of the state from a cued run into an uncued run transfers most of the cue benefit, while copying the fast half does not.
    \item \textbf{Recall by list position (Fig.~\ref{fig:app-position}):} across 256- and 1{,}024-record lists, cued recall stays roughly flat across positions while uncued recall drops for earlier records.
\end{enumerate}

\clearpage
\section{Pixel recall and compression tasks}
\label{app:exp-features}
\label{app:phase}
\label{app:stats}

This section provides experimental details and additional results for the inference-time compression experiments in Sec.~\ref{sec:features} (Figs.~\ref{fig:mnist-basic}--\ref{fig:phase-diagram}). This covers the MNIST/EMNIST pixel reconstruction task, the Hanzi character task and the letter-heat phase diagram experiments of the main text. Additionally, we include a sliding-window control experiment where a Transformer is trained with a limited attention window.

\subsection{Experimental details}
\label{app:pixel-setup}
\label{app:latent-setup}
\label{app:heat-setup}

\paragraph{MNIST and EMNIST pixel recall.}
We binarize $28\times28$ MNIST digits \citep{Lecun_1998} with $\mathbb{1}[\text{pixel}\ge128]$ and split the 60{,}000 training images into 55{,}000 train and 5{,}000 validation images. Each context is made by choosing a random subset of size $k\in\{196,392,588,784\}$ pixels (25--100\%) to reveal. Each pixel is then encoded as a 4-token tuple (\texttt{OBS}, \texttt{ROW\_$r$}, \texttt{COL\_$c$}, \texttt{PIXEL\_$v$}) using a 64-token vocabulary covering format tokens, row/column tokens and ink/background tokens.

Each episode is constructed as:
\begin{enumerate}
    \item \texttt{BOS}.
    \item The observed pixel tuples.
    \item \texttt{END\_CONTEXT}.
    \item Eight queries, each written as \texttt{QUERY ROW\_$r$ COL\_$c$ ANSWER}.
\end{enumerate}

We compute the next-token prediction loss only on the binary answers after each \texttt{ANSWER} position. \textbf{Importantly}, during training all eight queries ask for already-observed pixels, so the model is never trained to predict missing pixels.

We compare two models of the same size:
\begin{itemize}
    \item A 4-layer RWKV-7 \citep{peng2025rwkv7gooseexpressivedynamic} with $d=128$ and head size 16 (1.08M parameters, 9{,}216 state floats).
    \item A 4-layer Transformer with $d=128$, 2 heads, RoPE and Canon layers \citep{allenzhu2026physicslanguagemodels41}\footnote{The Transformer gets stuck during optimization without Canon layers.} (1.08M parameters, KV cache up to 3.2M floats).
\end{itemize}

We report \emph{balanced accuracy}, the mean of the accuracy on ink pixels and the accuracy on background pixels. This metric corrects for class imbalance (since most pixels are background, always predicting background scores 0.5).

As in App.~\ref{app:lm-eval}, each context is encoded into a state once and we query all missing pixels from that state to ``reconstruct'' the image (evaluated on 16 fixed validation images at 196, 392 and 588 observed pixels; 18{,}816 queries).

To verify that completion uses the given episode rather than just an average digit prior (Sec.~\ref{sec:mnist}), we also run an ink-matched control: we feed the observed pixels of a different digit with the most similar context and score against the original target digit's missing pixels.

For EMNIST \citep{Cohen_2017}, we use ten test images per class across all 52 uppercase and lowercase letter classes (196 observed pixels). We compare against a pixel-frequency baseline that predicts ink at any coordinate where more than half of MNIST training digits have ink (essentially the average MNIST digit). The ink/background split (Fig.~\ref{fig:app-emnist}c) and the galleries use five images each of eight letters (A--D, a--d).

\paragraph{Downstream classification.}
For digit classification, we use a special-token approach: after the recall-trained model reads the usual pixel context, we append a randomly initialized \texttt{CLASSIFY} token after \texttt{END\_CONTEXT} and project its final residual into a ten-way classification head. We fine-tune the full model to measure the data efficiency of learning the classes. We evaluate $\{1,2,4,\dots,2048,4950\}$ labels per class across three random seeds, with contexts of 196, 392 and 588 pixels, lr $10^{-4}$ and batch size 50, training for 2{,}000--10{,}000 steps with early stopping once validation saturates (8 validations without improvement). For EMNIST classification, we use the eight letters A--D and a--d (chance 0.125, up to 2{,}726 labels per class).

\paragraph{Hanzi characters.}
For the Hanzi character experiments, we render simplified-Chinese characters from Heiti SC Light on a $64\times64$ canvas, binarize them at the same 128 threshold and apply mild augmentations. We split by character into 930 training characters (512 instances each) and 163 held-out characters (256 instances each), constructed so that every component of a held-out character appears in at least three training characters. Each observed pixel is encoded as a 3-token tuple (\texttt{ROW COL PIXEL}, 134 tokens total) and 15--75\% of pixels are observed.

We train three models using the MNIST backbones:
\begin{itemize}
    \item RWKV-7.
    \item A full-attention Transformer with Canon layers.
    \item The KV-budget Transformer of App.~\ref{app:synthetic-setup} with $B=512$ tokens.
\end{itemize}

To read out whether the model state represents radicals, we fit a logistic ridge regression probe on the normalized 128-d \texttt{END\_CONTEXT} state given 15\% of pixels as context (614 pixels). We probe for eight different radicals and report macro AUC over the 163 held-out characters. For hidden-pixel completion at 3, 5, 8 and 15\% exposure, we compare against a Gaussian blur fitted per exposure and thresholded to match the true ink budget.

\paragraph{Letter-heat phase diagram.}
For this experiment, we construct each scene by taking an uppercase EMNIST letter \citep{Cohen_2017}, centering and blurring it ($\sigma=1.25$), scaling its peak by $\{4,6,8\}$ above a background of 1, adding Gaussian noise (s.d.\ 0.25) and quantizing readings to 16 integer values on a $24\times24$ grid (576 cells). We use disjoint EMNIST splits for train and test, and observe the same 192 cells in every scene (leaving 384 unseen).

In each episode, the query cell $Q$ is drawn uniformly among the 192 observed cells, and the cue cell $U$ is sampled from a distribution centered near $Q$ with spread $t$ (specifically row $Q$ of $\exp(-tL)$, where $L$ is the 4-nearest-neighbor graph Laplacian on observed cells). A larger $t$ spreads the cue further from the query; we train with $t\in\{0.25,0.375,0.5,1,4\}$. Sequences follow \texttt{BOS CUE} $x_U\,y_U$ \texttt{DOC} $(x\,y\,v)\times192$ \texttt{END\_DOC ASK} $x_Q\,y_Q$ \texttt{ANS} $v$ \texttt{EOS} (588 tokens, readings shuffled), with loss only on the observed answer token.

All phase-diagram models have 4 layers and width 256:
\begin{itemize}
    \item \textbf{Full attention:} A Transformer with Canon layers and 4 heads (4.29M params).
    \item \textbf{KV-budget Transformer:} The streaming token-dropping scheme of App.~\ref{app:synthetic-setup} with cross-attention to the cue and Gumbel noise $\tau=0.5$ during training, keeping $B\in[24,575]$ of the 576 evidence tokens (4.36M params).
    \item \textbf{RWKV-7:} 16 heads of size 16 (3.75M params, 16{,}384 floats). Smaller states are made by disabling $h$ heads and keeping $k$ key dimensions, giving $(16-h)\times k\times16\times4$ floats (256 to 16k).
\end{itemize}

Every run is evaluated on 48 test scenes with a common $t=4$ cue. We use two metrics:
\begin{itemize}
    \item \textbf{Unseen RMSE:} The root mean squared error on the 384 unseen cells between the predicted expected value and the rounded ground-truth field. Intervals are 95\% percentile bootstraps over seeds and scenes (2{,}000 resamples).
    \item \textbf{Legibility:} The top-1 accuracy of a frozen CNN letter classifier on the reconstructed maps (error bars: min--max over seeds). The classifier has 142{,}842 params and is trained only on clean renders, scoring 98.5\% on clean test renders and 93.8\% on interpolation maps.
\end{itemize}

\paragraph{Sliding-window control.}
Each of $Q=128$ keys has a fixed unit vector $a_q\in\mathbb{R}^3$. Every episode draws $z\sim\mathcal{N}(0,I_3)$ and sets the value of key $q$ to $\mathrm{QuantileBin}(a_q^\top z)$, one of $B=4$ bins. With probability $p$, this value is replaced with a uniform random draw. The best achievable accuracy is therefore $1-p+p/B$: at $p=0$ three numbers determine the whole episode, and at $p=1$ completion is at chance (0.25). Each context shows $M=32$ random keys as \texttt{BOS (KEY VALUE)$^{32}$ END\_CONTEXT} (66 tokens), followed by \texttt{QUERY KEY}. As usual, we only train on observed keys and score completion on the 96 missing keys.

We use the MNIST Transformer+Canon backbone and compare three ways of limiting memory:
\begin{itemize}
    \item \textbf{Window-trained:} Models trained with a strict sliding window of 60, 52, 44 or 36 tokens, leaving 28, 24, 20 or 16 records visible.
    \item \textbf{Bounded-window recomputation:} The full-cache model run at inference time with the same window.
    \item \textbf{Post-hoc deletion:} The full-cache model with old cache entries deleted at inference time.
\end{itemize}
We use five paired seeds, where each pair shares its initialization and batch stream. Runs at $p=1$ and $p=0.1$ use seed 0. Table~\ref{tab:pixel-hparams} summarizes the training hyperparameters for all four tasks in this section, and Table~\ref{tab:seeds-summary} lists the number of independent runs across all experiments in the paper.

\begin{table}[ht]
\centering
\footnotesize
\setlength{\tabcolsep}{4pt}
\caption{\textbf{Training hyperparameters for pixel recall, Hanzi, sliding-window control and phase-diagram tasks} (Sec.~\ref{sec:features}). All use AdamW \citep{loshchilov2019decoupled} with $\beta=(0.9,0.95)$, weight decay $0.01$, peak lr $3\times10^{-4}$, 2.5\% linear warmup and grad clip 1.0.}
\label{tab:pixel-hparams}
\begin{tabular}{lllll}
\toprule
Task & Backbone & Steps & Batch & Precision / Seeds \\
\midrule
MNIST / EMNIST pixels & 4L, $d=128$ (1.08M) & 15{,}000 & 64 & bf16 / 3 seeds (0--2) \\
Hanzi characters & 4L, $d=128$ (1.08M) & 100{,}000 & 16 & fp32 master / 3 seeds \\
Sliding-window control & 4L, $d=128$ (Transformer+Canon) & 5{,}000 & 512 & bf16 / 5 paired seeds \\
Letter-heat phase diagram & 4L, $d=256$ (3.75--4.36M) & 100{,}000 & 128 & bf16 / 3--5 seeds (300 runs) \\
\bottomrule
\end{tabular}
\end{table}

\begin{table}[ht]
\centering
\footnotesize
\caption{\textbf{Summary of independent runs across all main-text and appendix claims.}}
\label{tab:seeds-summary}
\begin{tabular}{p{0.48\linewidth}p{0.44\linewidth}}
\toprule
Experiment & Independent runs \\
\midrule
Pretrained language models (Figs.~\ref{fig:real-pretrained}, \ref{fig:app-roster}) & 1 evaluation pass, 8 independent profile sets \\
Fine-tuned language models (Figs.~\ref{fig:real-finetuned}, \ref{fig:app-ft-models}--\ref{fig:app-ft-short}) & 2 runs for World 2.9B / GDN / DeltaNet; 1 for others \\
Synthetic cued recall, main (Fig.~\ref{fig:synthetic}) & 1 training seed per model \\
Synthetic, state-matched sweep (Figs.~\ref{fig:app-acquisition}, \ref{fig:app-cue-write}, \ref{fig:app-position}) & 3 seeds per architecture (RWKV-7 $N\le256$: 9) \\
MNIST / EMNIST completion \& classification (Figs.~\ref{fig:mnist-basic}, \ref{fig:mnist-downstream}b) & 3 training seeds per architecture (3 label-subset seeds each) \\
Hanzi readout and completion (Figs.~\ref{fig:mnist-downstream}d, \ref{fig:app-hanzi}) & 3 training seeds per architecture \\
Sliding-window control (App.~\ref{app:window-pressure}) & 5 paired seeds \\
Letter-heat phase diagram (Figs.~\ref{fig:phase-diagram}, \ref{fig:app-kv}--\ref{fig:app-phase-archs}) & 3--5 seeds per setting (300 runs; GDN/DeltaNet: 3 seeds) \\
\bottomrule
\end{tabular}
\end{table}

\subsection{Additional results}
\label{app:gallery}

\paragraph{Pixel completion gallery and EMNIST OOD transfer.}
\label{app:emnist}

\begin{figure}[!htbp]
    \centering
    \includegraphics[width=0.9\textwidth]{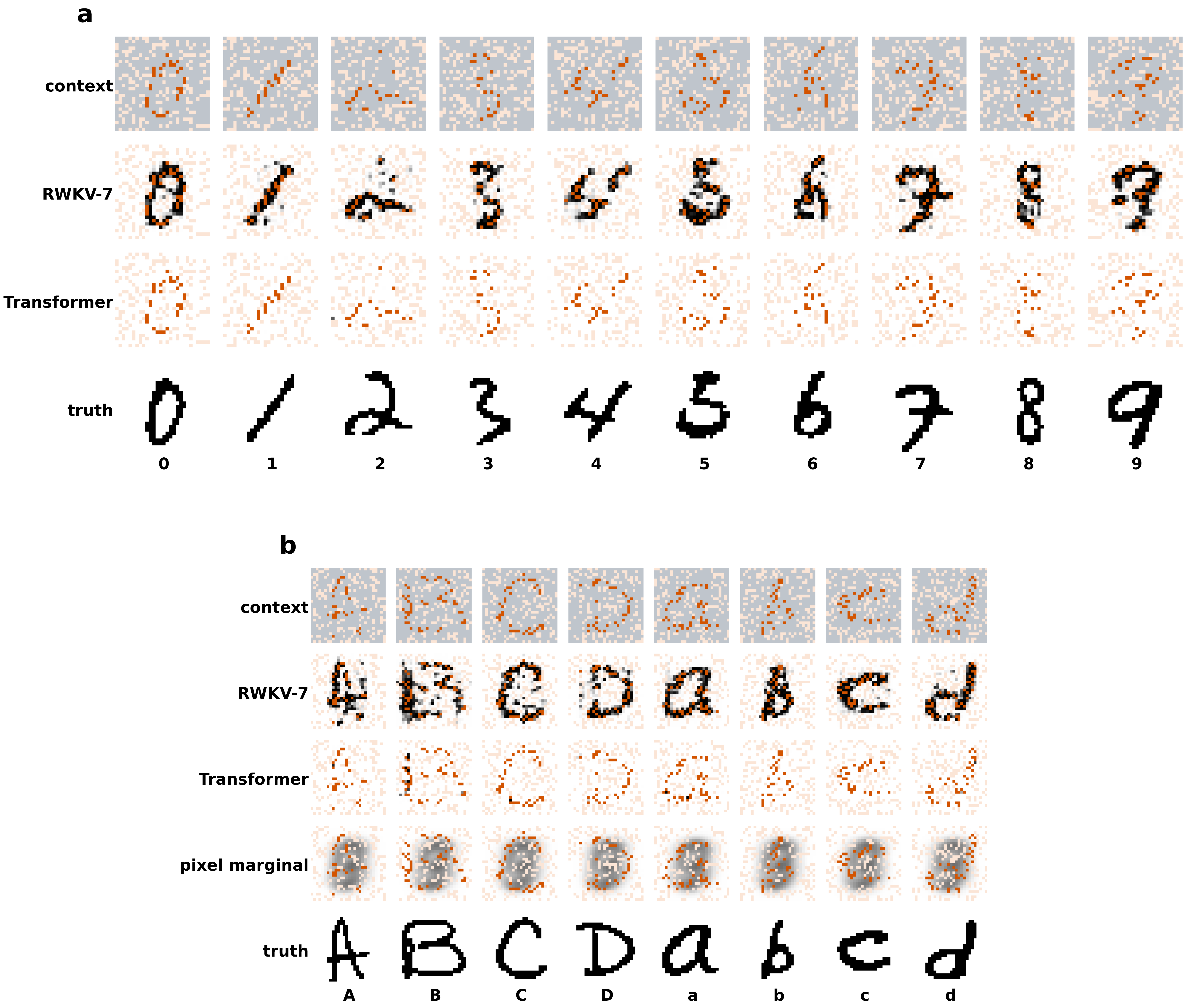}
    \caption{\textbf{Pixel completion gallery (196 observed pixels, MNIST-trained models).}
    \textbf{a.} One MNIST validation context per digit: the context, RWKV-7 and Transformer
    completions, and the true digit.
    \textbf{b.} The first test context of each of the eight EMNIST letters A--D, a--d, never
    seen in training, with the MNIST pixel-frequency map for comparison.
    Context row: observed ink (orange), observed background (pale), hidden (grey). Other rows:
    predicted ink probability at hidden pixels (black = 1), with the observed pixels overlaid.}
    \label{fig:app-gallery}
\end{figure}

Fig.~\ref{fig:app-gallery} shows example completions on MNIST digits and unseen EMNIST letters. We can clearly see that RWKV-7 reconstructs the digit while the Transformer puts ink only where the context already gave ink.

\begin{figure}[!htbp]
    \centering
    \includegraphics[width=0.9\textwidth]{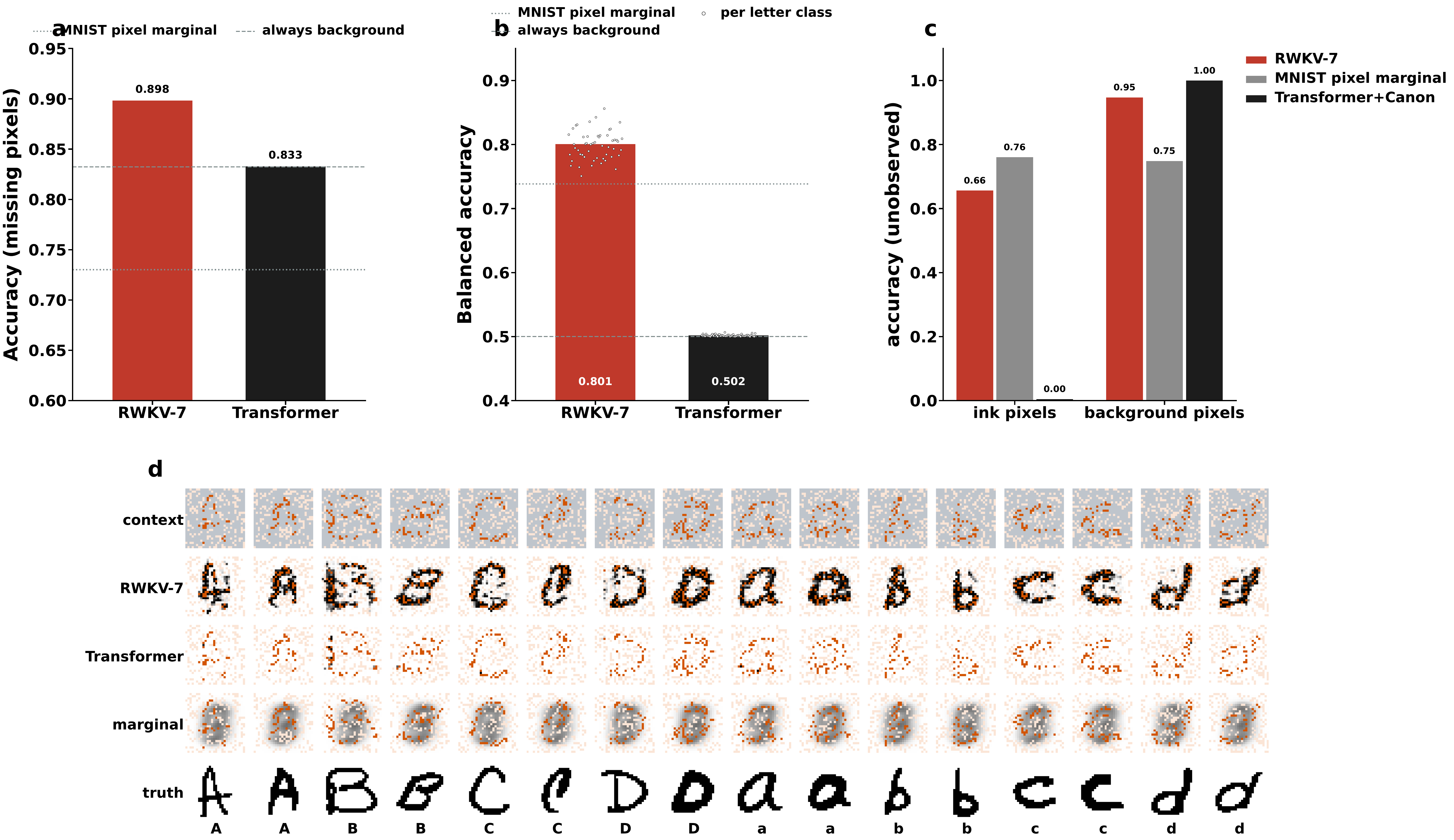}
    \caption{\textbf{OOD transfer to EMNIST letters (MNIST-trained, 196 observed pixels).}
    \textbf{a.} Raw accuracy on missing pixels over 520 contexts (52 classes $\times$ 10); dotted:
    MNIST pixel-frequency baseline; dashed: always predicting background.
    \textbf{b.} Balanced accuracy on the same contexts; dots: letter classes.
    \textbf{c.} Accuracy on hidden ink and hidden background pixels (eight letters A--D, a--d,
    five contexts each). The Transformer almost never predicts ink; RWKV-7 recovers two thirds
    of it while keeping the background clean.
    \textbf{d.} Two contexts per letter: context, RWKV-7, Transformer, pixel-frequency baseline and truth.}
    \label{fig:app-emnist}
\end{figure}

Fig.~\ref{fig:app-emnist} evaluates the MNIST-trained models on 52 EMNIST letter classes (10 contexts each) to see whether the model can roughly generate stroke-based writings out of distribution. Balanced accuracy (top panel) clearly separates the two models (0.801 for RWKV-7 vs.\ 0.502 for the Transformer on seed 0), and every RWKV-7 letter class (0.75--0.86) beats the MNIST pixel-frequency baseline (0.74).

\paragraph{Additional classification baselines.}
\label{app:input-matched}

\begin{figure}[!htbp]
    \centering
    \includegraphics[width=\textwidth]{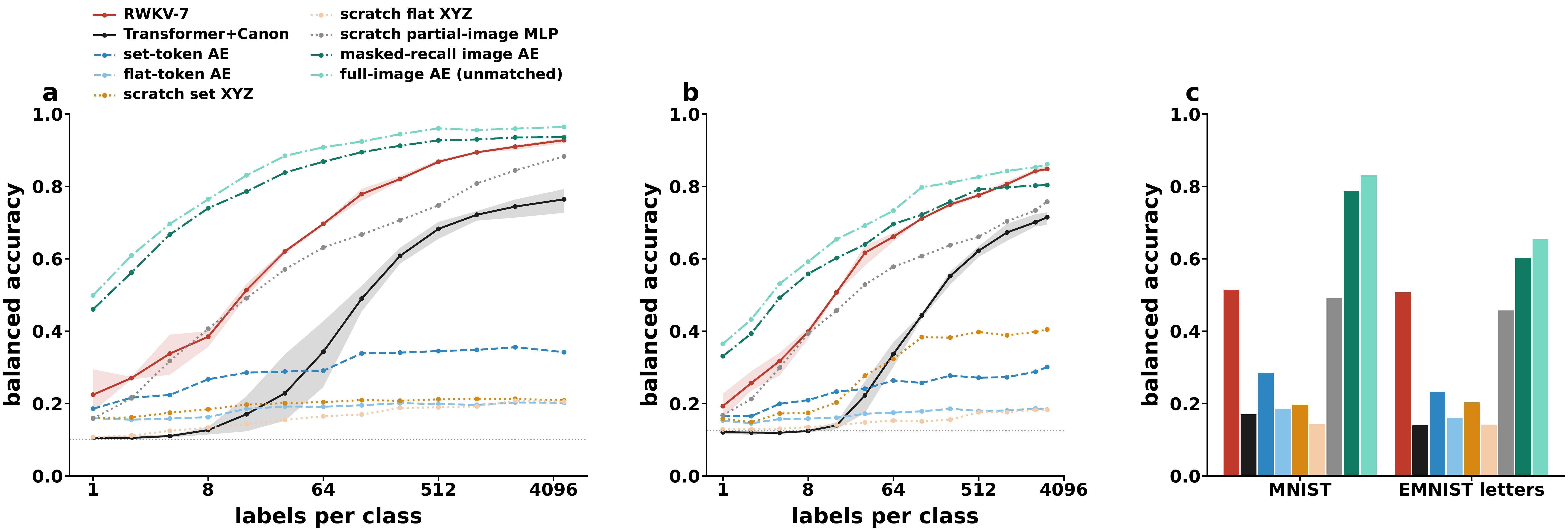}
    \caption{\textbf{Input-matched classification baselines.}
    \textbf{a, b.} Balanced accuracy against labels per class on MNIST (a) and the eight EMNIST
    letters A--D, a--d (b) for the recall-trained RWKV-7 and Transformer, autoencoders on the same
    records (set- and flat-token), from-scratch models on the records (set and flat $(x,y,z)$), a
    from-scratch MLP on the partial image, and image autoencoders (masked-recall and, as an unmatched
    upper control, full-image). Dotted: chance. The EMNIST sequence-model optimizers are not matched
    (App.~\ref{app:pixel-setup}).
    \textbf{c.} All arms at 16 labels per class, in the same order and colours.
    One pretraining seed per arm; shading: range over three label-subset seeds.}
    \label{fig:app-input-matched}
\end{figure}

Fig.~\ref{fig:app-input-matched} compares the classification heads of Fig.~\ref{fig:mnist-downstream}b against models that get the same pixels in other formats. Models that read the pixels as a list of records fall far behind RWKV-7 at 16 MNIST labels per class: token autoencoders score 0.19--0.29 and coordinate models trained from scratch score 0.14--0.20. Models that see the pixels on a 2-D grid do better:
\begin{itemize}
    \item A partial-image MLP trained from scratch matches RWKV-7 at 16 labels (0.49 vs.\ 0.47) but falls behind with more labels (0.75 vs.\ 0.86 at 512).
    \item An image autoencoder trained on masked recall beats RWKV-7 at all MNIST label counts (0.79 vs.\ 0.47 at 16), though RWKV-7 overtakes it on EMNIST from 2{,}048 labels per class.
    \item A full-image autoencoder gives an upper bound for the case where all pixels are visible.
\end{itemize}

\paragraph{Hanzi readout and completion.}
\label{app:hanzi}

\begin{figure}[!htbp]
    \centering
    \includegraphics[width=0.9\textwidth]{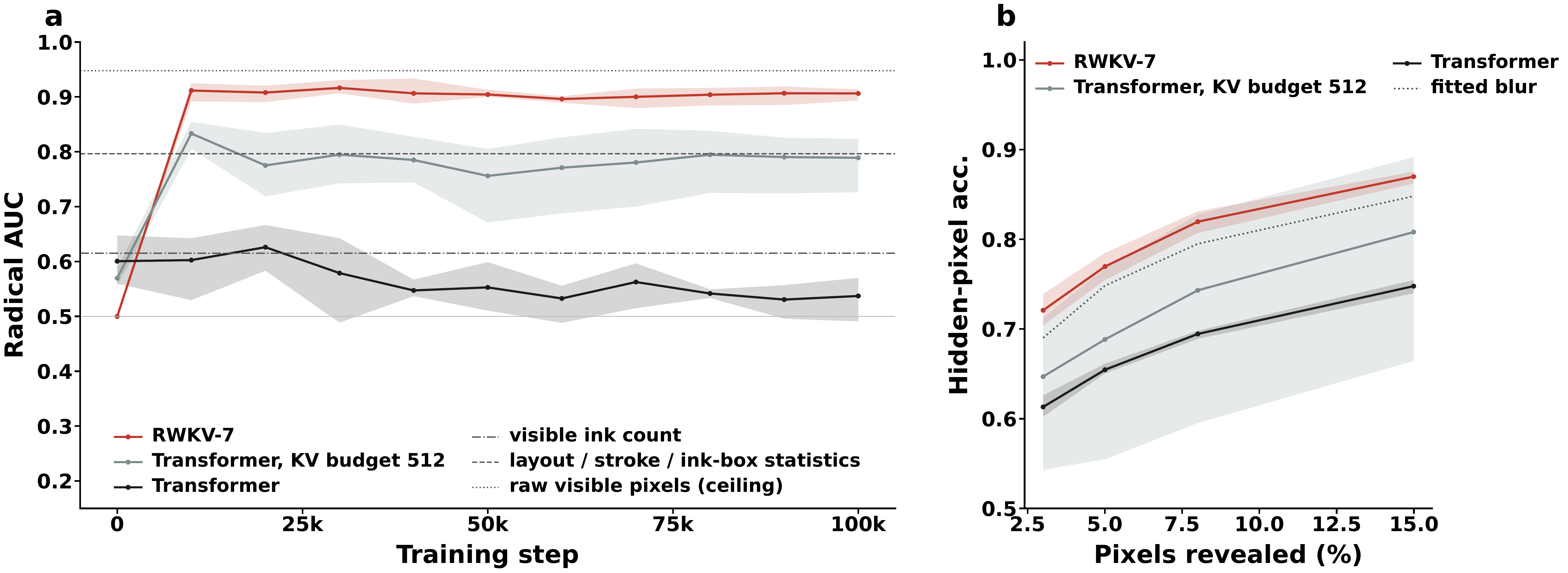}
    \caption{\textbf{Hanzi readout and completion (three seeds).}
    \textbf{a.} Radical macro AUC of a linear probe on the end-of-context state over training, for
    RWKV-7, a Transformer with a 512-token key--value budget and a full-attention Transformer.
    Horizontal lines (seed 0, no model): the same probe on image statistics of the visible pixels
    (ink count; layout, stroke and ink-box statistics) and on the raw visible pixels (ceiling).
    \textbf{b.} Balanced accuracy on hidden pixels of the 163 held-out characters against the
    fraction of pixels revealed; dotted: a Gaussian blur of the revealed ink, fitted per exposure.
    Lines: mean over seeds; shading: range over seeds.}
    \label{fig:app-hanzi}
\end{figure}

Fig.~\ref{fig:app-hanzi} tracks the radical readout of Fig.~\ref{fig:mnist-downstream}d over training. RWKV-7 reaches saturation of accuracy within 10k steps (0.89--0.91 across seeds at 100k). The 512-token KV-budget Transformer (0.73--0.82) sits close to the baseline using simple layout and stroke statistics of the visible pixels (0.80), while full attention (0.49--0.57) stays near chance. On hidden-pixel completion, RWKV-7 beats the fitted Gaussian blur at every exposure in all three seeds by 0.02--0.03. The budgeted Transformer matches RWKV-7 in two seeds (0.89 and 0.87 at 15\%) and fails in the third (0.66).

\paragraph{Attention-window pressure and MNIST attention pilots.}
\label{app:window-pressure}

As a third, smaller test of Sec.~\ref{sec:features}, we run the sliding-window control (App.~\ref{app:latent-setup}). We find that a Transformer trained with a strict sliding window (60, 52, 44 or 36 tokens) beats the full-cache Transformer of the same seed in 5/5 seeds at every window size, with median completion gains of $+0.166$, $+0.240$, $+0.284$ and $+0.292$. At 44 tokens, completion is $0.873$ (seed SD $0.005$) while recall of observed keys stays $>0.95$. Applying the window only at inference time, either by recomputation or post-hoc deletion, lowers completion in 5/5 seeds; these inference-only models also drop sharply at the edge of the visible window, while recall of the window-trained model stays smooth. When all values are replaced with uniform random noise ($p=1$, seed 0), all completion intervals include chance and recall drops for the window-limited models, showing that pressure cannot create predictability that is not in the data.

\begin{figure}[!htbp]
    \centering
    \includegraphics[width=\textwidth]{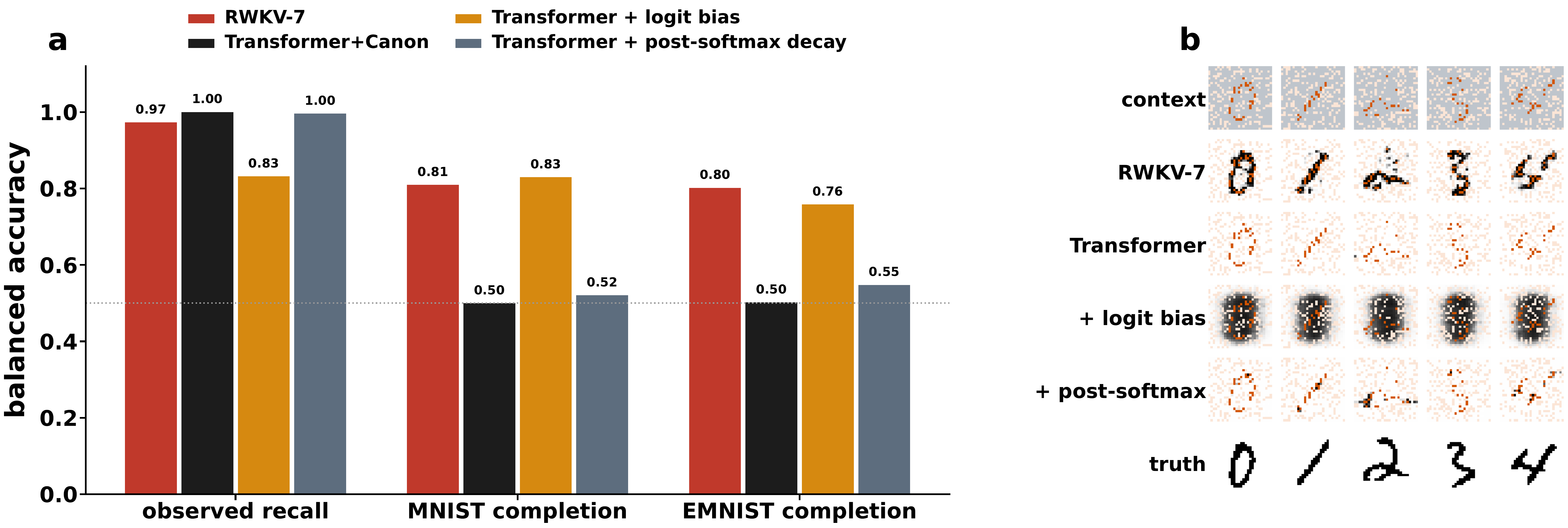}
    \caption{\textbf{Attention-pressure pilots in the MNIST Transformer.}
    \textbf{a.} Observed recall and missing-pixel completion (MNIST and EMNIST letters) for
    RWKV-7, the Transformer, and the Transformer with a logit bias or a post-softmax decay
    (512-token half-life); dotted: chance. Completion here uses a ten-context panel (RWKV-7
    0.81), not the 16-image suite of Fig.~\ref{fig:mnist-basic}.
    \textbf{b.} Five contexts: completions of each model and the true digit.
    One training seed per model.}
    \label{fig:app-pressure-pilots}
\end{figure}

Fig.~\ref{fig:app-pressure-pilots} tests two softer forms of pressure in the MNIST Transformer (one seed each, 512-token half-life):
\begin{itemize}
    \item \textbf{Normalized attention-logit bias:} Raises completion from 0.50 to 0.83 (0.76 on EMNIST), but drops observed recall to 0.83 and produces blurs similar to the pixel-frequency baseline.
    \item \textbf{Unnormalized decay after the softmax:} Keeps recall at 1.00 but does not recover completion (0.52).
\end{itemize}

\paragraph{Phase diagram: bounded KV-budget Transformers and mechanism.}
\label{app:phase-kv}
\label{app:phase-mechanism}

\begin{figure}[!htbp]
    \centering
    \includegraphics[width=\textwidth]{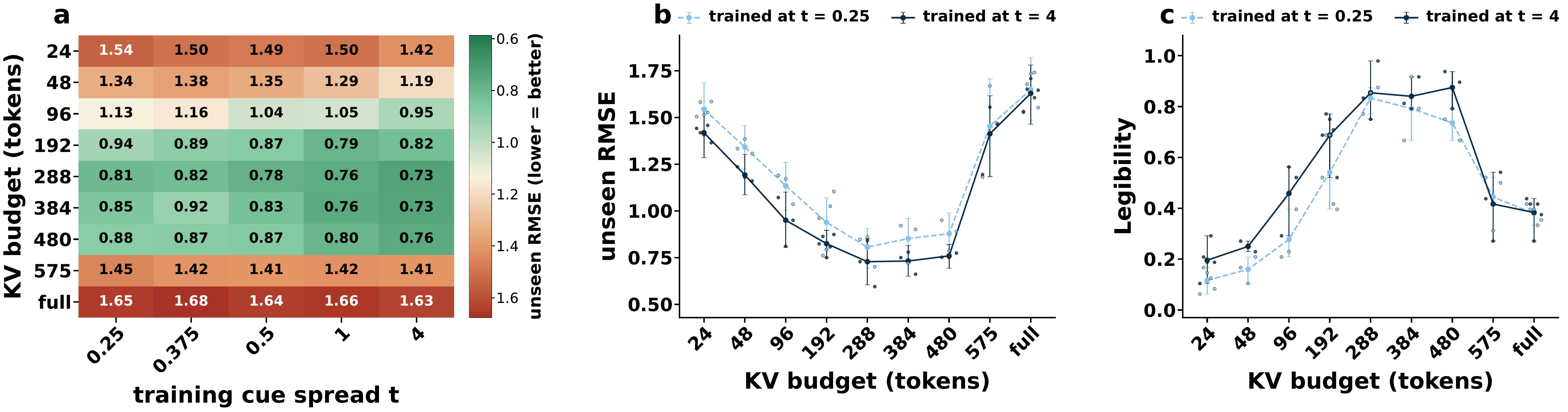}
    \caption{\textbf{Transformers with a bounded key--value budget}, as Fig.~\ref{fig:phase-diagram}c--e.
    \textbf{a.} RMSE on unseen cells by budget and training cue spread $t$ (full: unbounded
    attention); same colour scale as Fig.~\ref{fig:phase-diagram}c.
    \textbf{b.} RMSE on unseen cells for models trained at $t=0.25$ and $t=4$.
    \textbf{c.} Legibility, the accuracy of a frozen letter classifier on the reconstructed map.
    Dots: training seeds (3--5 per setting); error bars: 95\% bootstrap intervals over seeds and
    episodes (b) or range over seeds (c).}
    \label{fig:app-kv}
\end{figure}

Fig.~\ref{fig:app-kv} shows the KV-budget Transformer version of Fig.~\ref{fig:phase-diagram}c--e. Here, unseen RMSE is U-shaped with respect to the budget (the memory pressure): there is an optimal range (288--384 tokens, RMSE 0.73 at $t=4$) for generalizing to unseen cells. Given a budget that is too small, the model cannot store the letter (and tighter budgets are harder to train), while large budgets (1.41 at 575 tokens, close to full attention's 1.63) allow the model to fall back to the full-KV memorization solution. This experiment clearly shows that the gains come from the tight memory budget itself rather than from the token-dropping mechanism. Legibility, which we evaluate using the frozen CNN letter classifier, also shows a similar pattern (peaking at 288--480 tokens). The cue spread, here, has a relatively minor effect compared to the KV budget (changing unseen RMSE by at most 0.21).

\begin{figure}[!htbp]
    \centering
    \includegraphics[width=0.85\textwidth]{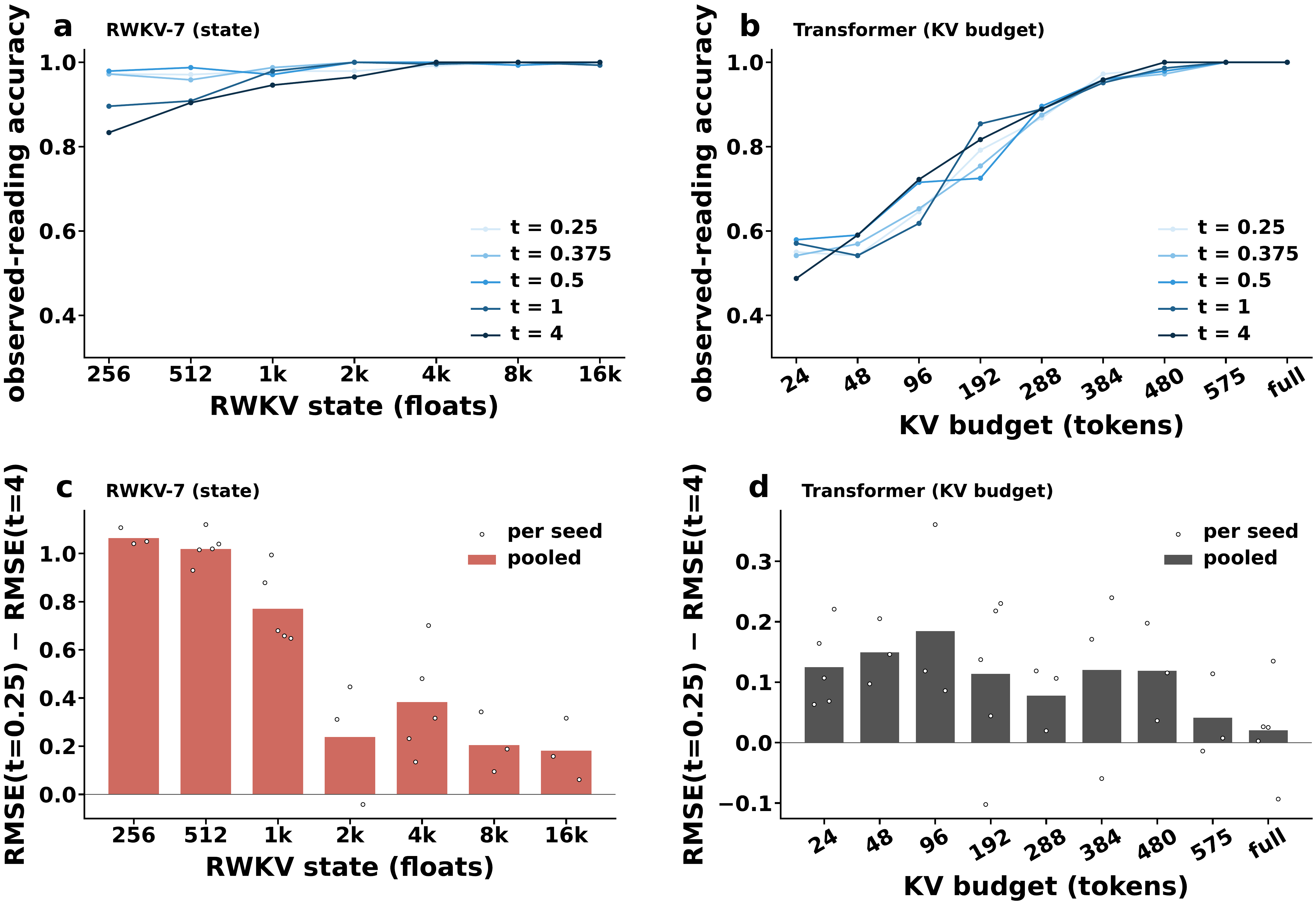}
    \caption{\textbf{How RWKV-7 and the budgeted Transformer respond differently to the cue.}
    \textbf{a, b.} Accuracy on observed readings (the training task, queried under each model's own
    training cue) by state size (a) or budget (b), one line per training cue spread $t$.
    \textbf{c, d.} Pressure effect, RMSE on unseen cells at $t=0.25$ minus at $t=4$; bars: pooled
    over seeds; dots: individual seeds.}
    \label{fig:app-phase-mechanism}
\end{figure}

\begin{figure}[!htbp]
    \centering
    \includegraphics[width=\textwidth]{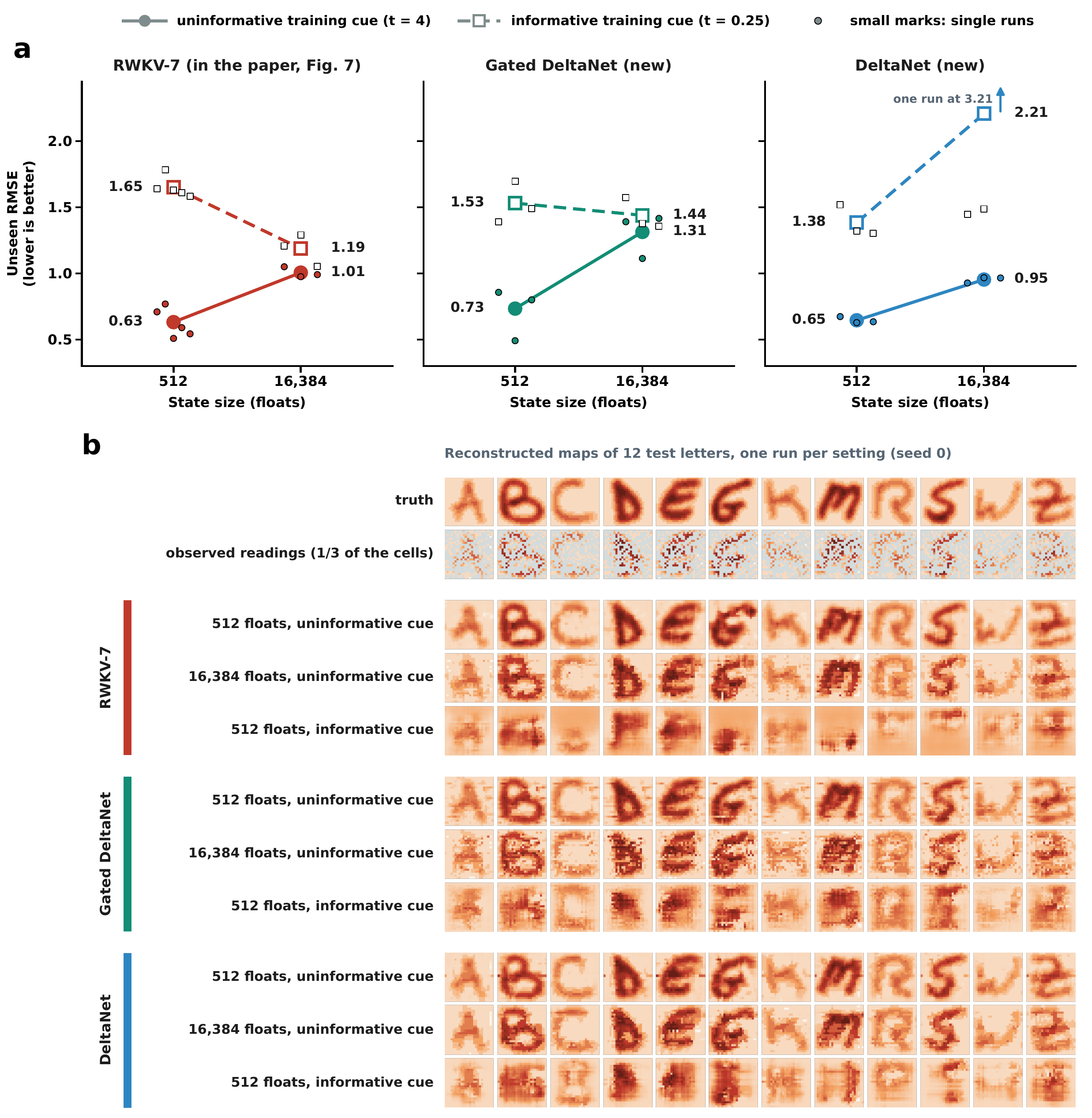}
    \caption{\textbf{The phase diagram of Sec.~\ref{sec:phase} in three recurrent architectures.} \textbf{a.} RMSE on unseen cells for models with a state of 512 or 16{,}384 floats, trained with an uninformative cue ($t=4$, solid) or an informative cue ($t=0.25$, dashed). Large marks: pooled over seeds. Small marks: single runs. \textbf{b.} Reconstructed maps of the 12 test letters of Fig.~\ref{fig:phase-diagram}b, one run per setting. Grey: cells that were not observed.}
    \label{fig:app-phase-archs}
\end{figure}

Fig.~\ref{fig:app-phase-mechanism} compares how RWKV-7 and the budgeted Transformer react to the cue. When given an informative cue ($t=0.25$), a small RWKV-7 state only needs to store the observed pixels near the cue, so it recalls almost every observed reading on the training task (0.97 at 256 floats) without needing to learn the letter. However, when given an uninformative cue ($t=4$), that shortcut disappears, causing observed recall to drop (0.83) and pushing the model to compress the whole letter instead (pressure effect $\approx 1$). For the Transformer, observed recall on the training task depends on the KV budget rather than the cue, keeping its pressure effect below 0.2 pooled over seeds. Finally, Fig.~\ref{fig:app-phase-archs} confirms that the same phase-diagram pattern holds for Gated DeltaNet and DeltaNet.

\end{document}